\documentclass{article}

\usepackage[toc,page,header]{appendix}

\usepackage{microtype}
\usepackage{graphicx}
\usepackage{subfigure}
\usepackage{booktabs} 

\usepackage{xcolor}
\definecolor{darkblue}{rgb}{0,0,0.5}
\definecolor{firebrick}{rgb}{0.75,0.125,0.125}
\definecolor{darkgreen}{rgb}{0,0.5,0}
\definecolor{light-gray}{gray}{0.5}

\usepackage[colorlinks=true,linkcolor=firebrick,citecolor=darkgreen,urlcolor=darkblue, pdfencoding=auto, psdextra]{hyperref}

\usepackage[accepted]{main}

\usepackage{amsmath}
\usepackage{amssymb}
\usepackage{mathtools}
\usepackage{amsthm}

\usepackage[capitalize,noabbrev]{cleveref}

\theoremstyle{plain}
\newtheorem{theorem}{Theorem}[section]
\newtheorem{proposition}[theorem]{Proposition}

\theoremstyle{definition}
\newtheorem{definition}[theorem]{Definition}

\theoremstyle{remark}
\newtheorem{remark}[theorem]{Remark}

\usepackage[textsize=tiny]{todonotes}

\usepackage{custom_notations}
\usepackage{multirow}
\usepackage{arydshln}
\usepackage{changes}
\usepackage{caption}
\usepackage{subcaption}
\usepackage{lipsum}
\usepackage{wrapfig}
\usepackage{minitoc}
\icmltitlerunning{}

\definecolor{darkbrown}{rgb}{0.4, 0.26, 0.13}

\definecolor{color-giuseppe}{rgb}{0.2, .8, 0.2}
\definechangesauthor[name={Giuseppe}, color=color-giuseppe]{G}

\definechangesauthor[name={Maurizio}, color=blue]{M}

\definechangesauthor[name={Albert}, color=orange]{A}

\definechangesauthor[name={Balazs}, color=cyan]{B}

\begin{document}
\doparttoc 
\faketableofcontents 


\twocolumn[
\icmltitle{A Multi-step Loss Function for Robust Learning of the Dynamics in Model-based Reinforcement Learning}



\icmlsetsymbol{equal}{*}

\begin{icmlauthorlist}
\icmlauthor{Abdelhakim Benechehab}{huawei,eurecom}
\icmlauthor{Albert Thomas}{huawei}
\icmlauthor{Giuseppe Paolo}{huawei}
\icmlauthor{Maurizio Filippone}{kaust}
\icmlauthor{Bal\'{a}zs K\'{e}gl}{huawei}
\end{icmlauthorlist}

\icmlaffiliation{huawei}{Noah's Ark Lab, Huawei Technologies France}
\icmlaffiliation{eurecom}{Department of Data Science, EURECOM, France}
\icmlaffiliation{kaust}{Statistics Program, KAUST, Saudi Arabia}

\icmlcorrespondingauthor{Abdelhakim Benechehab}{abdelhakim.benechehab1@huawei.com}

\icmlkeywords{Machine Learning, ICML}

\vskip 0.3in
]



\printAffiliationsAndNotice{}  

\begin{abstract}
In model-based reinforcement learning, most algorithms rely on simulating trajectories from one-step models of the dynamics learned on data. A critical challenge of this approach is the compounding of one-step prediction errors as the length of the trajectory grows. In this paper we tackle this issue by using a multi-step objective to train one-step models. Our objective is a weighted sum of the mean squared error (MSE) loss at various future horizons. We find that this new loss is particularly useful when the data is noisy (additive Gaussian noise in the observations), which is often the case in real-life environments. To support the multi-step loss, first we study its properties in two tractable cases: i) uni-dimensional linear system, and ii) two-parameter non-linear system. Second, we show in a variety of tasks (environments or datasets) that the models learned with this loss achieve a significant improvement in terms of the averaged R2-score on future prediction horizons. Finally, in the pure batch reinforcement learning setting, we demonstrate that one-step models serve as strong baselines when dynamics are deterministic, while multi-step models would be more advantageous in the presence of noise, highlighting the potential of our approach in real-world applications.
\end{abstract}

\section{Introduction}
\label{sec:introduction}
In reinforcement learning (RL) we learn a control agent (or policy) by interacting with a dynamic system (or environment), receiving feedback in the form of rewards. This approach has proven successful in addressing some of the most challenging problems, as evidenced by \citet{SilveR2017, SilveR2018, Mnih2015, Vinyals2019}. However, RL remains largely confined to simulated environments and does not extend to real-world engineering systems. This limitation is, in large part, due to the scarcity of data resulting from operational constraints (physical systems with fixed time steps or safety concerns). Model-based reinforcement learning (MBRL), can potentially narrow the gap between RL and applications thanks to a better sample efficiency.

\begin{figure}[t]
\begin{center}
\centerline{\includegraphics[width=0.95\columnwidth]{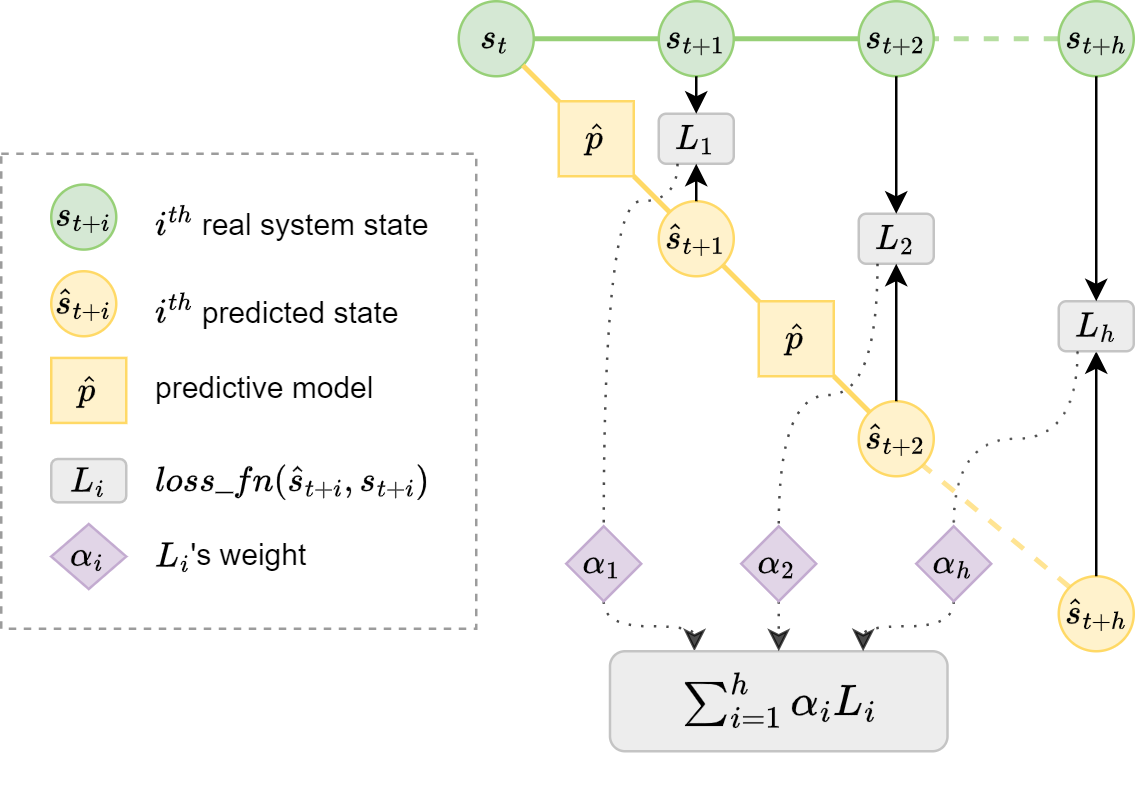}}
\caption{Schematic representation of the multi-step prediction framework using a one-step predictive model $\hat{p}$. The diagram illustrates the iterative prediction of future states $\hat{s}_{t+i}$, the computation of per-horizon losses $L_i$ against real system states $s_{t+1}$, and the weighting of these losses $\alpha_i$ to optimize the predictive model over a horizon of $h$ steps.}
\label{fig:diagram}
\end{center}
\vskip -0.3in
\end{figure}

MBRL algorithms alternate between two steps: i) model learning, a supervised learning problem to learn the dynamics of the environment, and ii) policy optimization, where a policy and/or a value function is updated by sampling from the learned dynamics. MBRL is recognized for its sample efficiency, as policy/value learning is conducted (either wholly or partially) from imaginary model rollouts (also referred to as background planning), which are more cost-effective and readily available than rollouts in the true dynamics \citep{Janner2019}. Moreover, a predictive model that performs well out-of-distribution facilitates easy transfer to new tasks or areas not included in the model training dataset \citep{Yu2020}.

While MBRL algorithms have achieved significant success, they are prone to \emph{compounding errors} when planning over extended horizons \citep{Lambert_2022}. This issue arises due to the propagation of one-step errors, leading to highly inaccurate predictions at longer horizons. This can be problematic in real-world applications, as it can result in out-of-distribution states that may violate the physical constraints of the environment, and misguide planning. The root of compounding errors is in the nature of the models used in MBRL. Typically, these are one-step models that predict the next state based on the current state and executed action. Long rollouts are then generated by iteratively applying these models, leading to compounding errors. To tackle this problem, our approach involves adjusting the training objective of these models to focus on optimizing for long-horizon error (\cref{fig:diagram}). This strategy is especially beneficial in presence of additive noise in observations, a context that mirrors real-world scenarios. Indeed, in situations where state perturbations occur due to unavoidable measurement errors or adversarial attacks \citep{Zhang2021, Sun2023}, aiming to minimize the long-horizon error is advantageous. It enhances the signal-to-noise ratio by incorporating future information, thereby improving the practical effectiveness of the models.


Our key contributions are the following.

\begin{itemize}
    \item We propose a novel training objective for one-step predictive models consisting in a weighted sum of MSE (Mean Squared Error) losses at several future horizons.
    \item We use two tractable cases to demonstrate the advantages of our weighted multi-step loss. In a linear system, this loss allows for the identification of minimizers achieving a bias-variance trade-off. In the case of a two-parameter neural network, we show how important the weights are to achieve strong performance across multiple future horizons.
    \item Finally, we analyze the optimal weight configurations of the multi-step loss and show that models trained with this loss improves over the one-step baseline across diverse environments, datasets, and levels of noise.
\end{itemize}

\section{Preliminaries}
\label{sec:Preliminaries}

The conventional framework used in RL is the finite-horizon \textbf{Markov decision process (MDP)}, defined by the tuple $\mathcal{M} = \langle  \mathcal{S}, \mathcal{A}, p, r, \rho_0, \gamma \rangle$. 
In this notation, $\mathcal{S}$ is the state space and $\mathcal{A}$ is the action space. 
The transition dynamics, which could be stochastic, are represented by $p : \mathcal{S} \times \mathcal{A} \leadsto \mathcal{S}$\footnotemark. The reward function is denoted by $r : \mathcal{S} \times \mathcal{A} \rightarrow \mathbb{R}$. The initial state distribution is given by $\rho_0$, and the discount factor is represented by $\gamma \in (0,1]$. The objective of RL is to identify a policy $\pi : \mathcal{S} \leadsto \mathcal{A}$. This policy aims at maximizing the expected sum of discounted rewards, denoted as $J(\pi, \mathcal{M}) := \mathbb{E}_{s_0 \sim \rho_0, a_t \sim \pi, \, s_{t>0} \sim p}[ \sum_{t=0}^{H} \gamma^t r(s_t, a_t) ]$, where $H$ represents the horizon of the MDP.

\footnotetext{we use $\leadsto$ to denote both probabilistic and deterministic mapping.}


\textbf{Model-based RL (MBRL)} algorithms learn a model of the dynamics of the environment $\hat{p}$ (and sometimes of the reward function $\hat{r}$ as well) in a supervised fashion from data collected when interacting with the real system $p$. The loss function is typically the negative log-likelihood $\mathcal{L}(\mathcal{D}; \hat{p}) = - \frac{1}{N} \sum_{i=1}^N \log  \hat{p}(s^i_{t+1}|s^i_t, a^i_t)$ for stochastic models or mean squared error (MSE) in the case of deterministic ones.
The learned model can subsequently be employed for policy search under the MDP $\widehat{\mathcal{M}} = \langle  \mathcal{S}, \mathcal{A}, \hat{p}, r, \rho_0, \gamma \rangle$.
This MDP shares the state and action spaces $\mathcal{S}, \mathcal{A}$, and the reward function $r$, with the true MDP $\mathcal{M}$, but uses the dynamics $\hat{p}$ learned from the dataset $\mathcal{D}$. The policy $\hat{\pi} = \argmax_\pi J(\pi, \widehat{\mathcal{M}})$ learned on $\widehat{\mathcal{M}}$ is not guaranteed to be optimal under $\mathcal{M}$ due to distribution shift and model error.


In this work, we also consider state-perturbed MDPs that fall under the formalism of  \textbf{partially-observed Markov decision process (POMDP)}. Indeed, a POMDP is defined by the underlying MDP $\langle  \mathcal{S}, \mathcal{A}, p, r, \rho_0, \gamma \rangle$, in addition to a set of possible observations $\Omega$, and an observation function $f_o : \mathcal{S} \times \mathcal{A} \leadsto \Omega$ ( where $o_t \sim f_o(\cdot|a_{t-1},s_t)$). In practice, we consider systems with additive Gaussian noise in observations $o_t = s_t + \epsilon$ where $\epsilon \sim \cN(0,\Sigma)$, which is a special case of a POMDP. This formalism is closely related to \emph{state-noisy} MDPs \citep{Sun2023} and \emph{state-adversarial} MDPs \citep{Zhang2021} where an attacker engineers the perturbation to intentionally decrease the performance of solvers.


The formal definition of the multi-step loss is given in \cref{sec:multistep}. We then present the related work in \cref{sec:Related}. We study the properties of the multi-step loss in \cref{sec:Understanding}. This is done through two tractable cases: a uni-dimensional linear system in and a two-parameter non-linear system. Finally, the experimental setup and the performance evaluation of the models is discussed in Section~\ref{sec:Experiments}.

\section{The multi-step loss}
\label{sec:multistep}

In MBRL, it is common to use a parametric model $\hat{p}_\theta$ that predicts the state\footnotemark{} one-step ahead $\hat{s}_{t+1} \rLeadsto \hat{p}_\theta(s_t, a_t)$.
We train this model to optimize the one-step predictive error $L\big(s_{t+1}, \hat{p}_\theta(s_t, a_t)\big)$ (MSE or NLL for stochastic modeling) in a supervised learning setting. 
To learn a policy, we use these models for predicting $h$ steps ahead by applying a procedure called \emph{rollout}.

\footnotetext{In this section, we do not make the distinction between states $s$ and observations $o$ as the definition of the multi-step loss is independent of the underlying MDP.}

\begin{definition}
\label{def:rollout}
(Rollout). We generate $\hat{s}_{t+j} \rLeadsto \hat{p}_\theta(\hat{s}_{t+j-1}, a_{t+j-1})$ recursively for $j = 1, \ldots, h$, to collect a trajectory $\tau = (s_0, a_0, \hat{s}_{1}, a_{1}, ..., \hat{s}_{j}, a_{j}, ...)$,
where $(a_t, \ldots, a_{t+h-1}) = \ba_{t:t+h-1} = \ba_{\tau}$ is either a fixed action sequence generated by planning or sampled from a policy $a_{t+j} \rLeadsto \pi(s_{t+j})$ for $j = 1, \ldots, h$, on the fly.
\end{definition}

Formally, let

\vskip -0.24in
\begin{align*} 
\hat{s}_{t} &= s_t \\
\hat{s}_{t+j-1} &\rLeadsto \hat{p}_\theta(\hat{s}_{t+j-2}, a_{t+j-2}) \\
\hat{p}_\theta^j(s_t, \ba_{t:t+j-1}) &= \hat{p}_\theta\big(\hat{s}_{t+j-1}, a_{t+j-1}\big) \hspace{0.2cm} \text{ for } j = 1, \ldots, h \\
\end{align*}
\vskip -0.1in

Using $\hat{p}_\theta^h(s_t, \ba_{t:t+h-1})$ to estimate $s_{t+h}$
is problematic for two reasons:
\begin{itemize}
\item A distribution mismatch occurs between the inputs that the model was trained on ($s_{t+1} \rLeadsto p(s_t, a_t)$) and the inputs the model is being evaluated on ($\hat{s}_{t+1} \rLeadsto \hat{p}_\theta(s_t, a_t)$\cite{Talvitie_2014, Talvitie_2017}.
\item The predictive error (and the modeled uncertainty in the case of stochastic models) will propagate through the successive model calls, leading to compounding errors \citep{Lambert_2022, Talvitie_2014, Venkatraman_2015}.
\end{itemize}

To mitigate these issues, we study models that, given the full action sequence $\ba_{t:t+h-1}$, learn to predict the state $s_{t+h}$ by recursively predicting the intermediate states $s_{t+j}$, for $j=1,\hdots,h$. We address this problem through the use of a weighted multi-step loss that accounts for the predictive error at different future horizons. 

\begin{definition}
\label{def:loss}
(Weighted multi-step loss). Given horizon-dependent weights $\balpha = (\alpha_1, \ldots, \alpha_h)$ with $\sum_{i=1}^h \alpha_i = 1$, a one-step loss function $L$, an initial state $s_t$, an action sequence $\ba_{\tau} = \ba_{t:t+h-1}$, and the real (ground truth) visited states $\bs_{\tau} = \bs_{t+1:t+h}$, we define the weighted multi-step loss as
\begin{equation*}
    L^h_\balpha\big(\bs_{\tau},\hat{p}_\theta(s_t, \ba_{\tau})\big) = \\ \sum_{j=1}^{h} \alpha_j L\big(s_{t+j},\hat{p}_\theta^j(s_t, \ba_{t:t+j-1})\big).
\end{equation*}
\end{definition}
The dependency of $L^h_\balpha$ on $h$ is omitted in the rest of the paper when $h$ is clear from the context. Furthermore, the loss $L$ used in the multi-step loss $L_\balpha$ will always be the MSE.

\cref{alg:training} shows the training procedure of multi-step models. We emphasize that unlike popular methods in the existing literature (teacher forcing \citep{Williams1989}, hallucinated replay \cite{Talvitie_2014} and data-as-demonstrator \cite{Venkatraman_2015}), which consists in augmenting the training data with predicted states, our method consists in back-propagating the gradient of the loss through the successive compositions of the model $\hat{p}_\theta$ figuring in $L_\balpha$.

\begin{algorithm}[htb]
   \caption{Training a model using the multi-step loss}
   \label{alg:training}
\begin{algorithmic}
   \STATE {\bfseries Input:} model $\hat{p}_\theta$, trajectory $\tau=(s_{t+i},a_{t+i})_{i=0}^h$, loss function $L$, horizon $h$, normalized weights $\{\alpha_i\}_{i=1}^h$
   \STATE Initialize the loss $l = 0$
   \STATE $s=s_t$
   \FOR{$i=0$ {\bfseries to} $h-1$}
   \STATE Sample from the model $\hat{s}_{t+i+1} \rLeadsto \hat{p}_\theta(s, a_{t+i})$
   \STATE Update the loss $l \mathrel{+}= \alpha_i \cdot L(\hat{s}_{t+i+1}, s_{t+i+1})$
   \STATE $s=\hat{s}_{t+i+1}$
   \ENDFOR
   \STATE Update $\theta$ using a gradient step minimizing $l$
\end{algorithmic}
\end{algorithm}

\section{Related Work}
\label{sec:Related}


The premises of multi-step dynamics modeling can be tracked back to early work about temporal abstraction \citep{Sutton1999, Precup1998} and mixture of timescale models in tabular MDPs \citep{Precup1997, Singh1992, Sutton1985, sutton1995}. These works study fixed-horizon models that learn an abstract dynamics mapping from initial states to the states $j$ steps ahead. A different approach consists in optimizing the multi-step prediction error of single-step models that are used recursively, which we study here. This approach has been studied for recurrent neural networks (RNNs) and is referred to as \emph{teacher forcing} \citep{Lamb2016, Bengio2015, Huszar2015, Pineda1988, Williams1989}. The idea consists in augmenting the training data with predicted states.  More recent works have built on this idea \citep{Abbeel2005,Talvitie_2014, Talvitie_2017,Venkatraman_2015}. These methods, albeit optimizing for future prediction errors, assume the independence of the intermediate predictions on the model parameters, making them more of a data augmentation technique than a proper optimization of multi-step errors.



The closest works to ours which also consider the intermediate predictions to be dependent on the model parameters are \citet{Lutter2021}, \citet{Byravan_2021} and \citet{Xu2018}. These works all use an equally-weighted multi-step loss whereas we emphasize on the need of having a weighted multi-step loss. \citet{Nagabandi2018} only use an equally-weighted multi-step loss for validation, which is also common in the time series literature \citep{Tanaka1995, Fraedrich1998, McNames2002, Bentaieb_2012a, Bentaieb_2012b, Chandra_2021}. \citet{Lutter2021} and \citet{Xu2018} find that only small horizons ($h=2,3,5$) yield an improvement over the baseline, which we suggest is due to using equal weights in the multi-step loss. \citet{Byravan_2021} successfully use $h=10$ with equal weights in the context of model-predictive control (MPC). However, they consider it as a fixed design choice and might have tailored their approach accordingly. To the best of our knowledge we are the first work highlighting the importance of the weighting mechanism in balancing the multi-step errors. Futhermore, we particularly stress the importance of having a weighted multi-step loss for state-noisy MDPs and study the impact of the noise level. We believed that such a setting is still under-studied event though previous model-free methods have been proposed \citep{Hess2023, Sun2023, Zhang2021, Pattanaik2017} and a very recent work \citep{Noauthor2023} presents a model-based approach using diffusion models.



\section{Understanding the multi-step loss: two case studies}
\label{sec:Understanding}

\subsection{Uni-dimensional linear system}
\label{subsec:Uni}

The first case consists in studying the solutions of the multi-step loss for $h=2$ in the case of an uncontrolled linear (discrete) dynamical system with additive Gaussian noise, and a linear model. With such a simple formulation, we benefit from the fact that the optimization is tractable and closed-form solutions can be obtained for each $\balpha \in [0, 1]$. We start by defining the system and the model.

\begin{definition}
\label{def:uni}
(Uni-dimensional linear system with additive Gaussian noise). For an initial state $s_0 \in \R$ and an unknown parameter $\theta_{true} \in (-1,1)$ (for stability) we define the transition function and observations as
\begin{align*} 
s_{t+1} &= \theta_{true} \cdot s_t \\ 
o_{t+1} &= s_{t+1} + \epsilon_{t+1} \text{ with } \epsilon_{t+1} \sim \cN(0, \sigma^2) \text{ and } \sigma \in \R
\end{align*}
\end{definition}

\begin{definition}
\label{def:model_uni}
(Linear model). For an initial state $s_0 \in \R$ and a parameter $\theta \in \R$ that we learn by minimizing the multi-step loss for $h=2$, we define the linear model as
$\hat{s}_{t+1} = \hat{p}_\theta(s_t) = \theta \cdot s_t$
\end{definition}

In this setup, the multi-step loss for $h=2$ boils down to a polynomial in the model's parameter $\theta$:

\vskip -0.2in
$$
L_\alpha\big(\bo_{\tau},\hat{p}_\theta(s_t)\big) = \\ \alpha (\theta s_t - o_{t+1})^2 + (1 - \alpha) (\theta^2 s_t - o_{t+2})^2
$$
where $\bo_{\tau} = (o_{t+1}, o_{t+2})$. The aim of our study is to analyze the statistical properties of $\hat{\theta}(\alpha) \in \argmin_\theta L_\alpha\big(\bo_{\tau},\hat{p}_\theta(s_t)\big)$ for different values of $\alpha$ and different values of the observational noise scale\footnote{\emph{noise} and \emph{noise scale} are used interchangeably. In practice, $\sigma$ is computed as a percentage (e.g. $2\%$) of the state space width.} $\sigma$. 

\begin{figure}[ht]
\begin{center}
\centerline{\includegraphics[width=0.8\columnwidth]{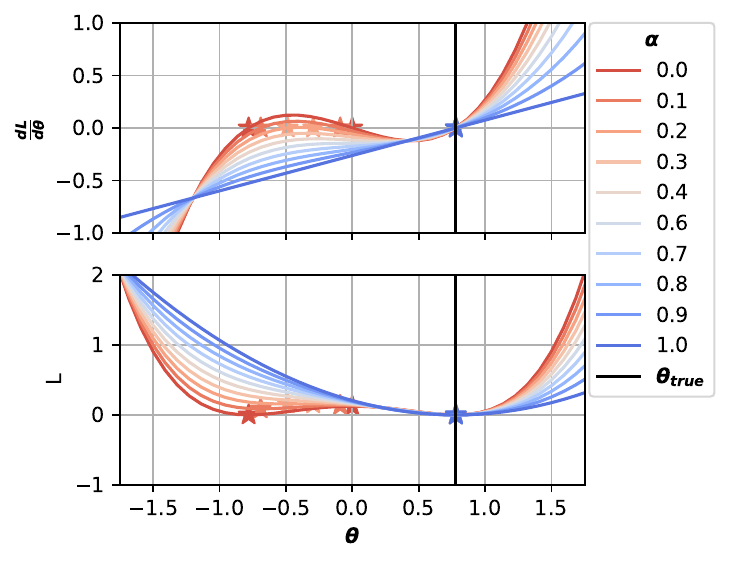}}
\vskip -0.1in
\caption{The loss function and its derivative for different values of $\theta$ and $\alpha$, in absence of noise ($\sigma=0$). In this figure, $\theta_{true}$ is fixed to a randomly selected value, $\theta_{true}=0.78$. The roots of the derivative are highlighted with stars.}  
\label{fig:l_and_dl}
\end{center}
\vskip -0.2in
\end{figure}

\cref{fig:l_and_dl} shows the loss function curve and its critical points for different values of $\alpha$. The minimizers $\hat{\theta}(\alpha)$ can be obtained by solving the polynomial equation $d L_\alpha/d \theta = 0$. When $\alpha \in (0,1)$, we compute the roots of the cubic polynomial equation using Cardano's formulas. These latter include at least one real root ($\alpha \geq 0.3$ in \cref{fig:l_and_dl}) and two (potentially real) conjugate complex roots ($\alpha < 0.3$ in \cref{fig:l_and_dl}).  

In the rest of the experiments, we fix a dataset of initial states $\cS_0$ and sample $K$ times a two-step transition, yielding a dataset $\cD = {(\cS_0, \cO^j_1, \cO^j_2)_{j=1,\hdots,K}}$. This dataset showcases different realizations of the observational noise, which is sampled i.i.d. from a Gaussian distribution $\cN(0, \sigma^2)$.

\begin{figure}[ht]
\begin{center}
\centerline{\includegraphics[width=0.8\columnwidth]{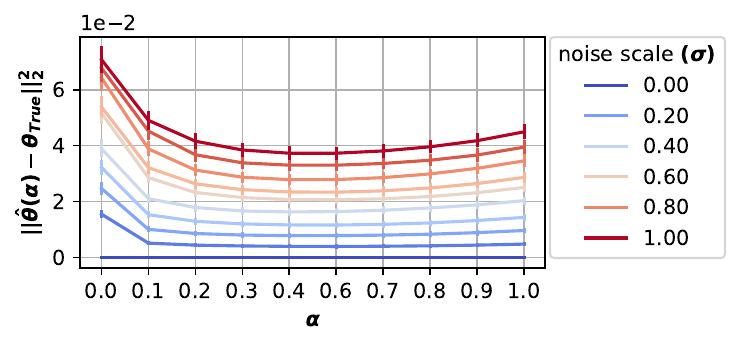}}
\vskip -0.1in
\caption{The distance between the true parameters and the optimal parameters for different values of $\alpha$ and noise scales.}
\label{fig:error_noise}
\end{center}
\vskip -0.2in
\end{figure}

We then compare the distance between the minimizers of the loss function with different values of $\alpha$ and the true parameter $\theta_{true}$. When there is more than one root, we assume access to the sign of $\theta_{true}$ so that we can choose the correct estimator $\hat{\theta}(\alpha)$. 
\cref{fig:error_noise} shows that as the noise increases, the vanilla MSE loss estimator ($\alpha=1$) is not the best estimator with respect to the distance to the true parameter $\theta_{true}$. 
Interestingly, the best solution is obtained for $\alpha \in (0, 1)$.


\begin{figure*}[ht]
\centering
\includegraphics[width=1.0\textwidth]{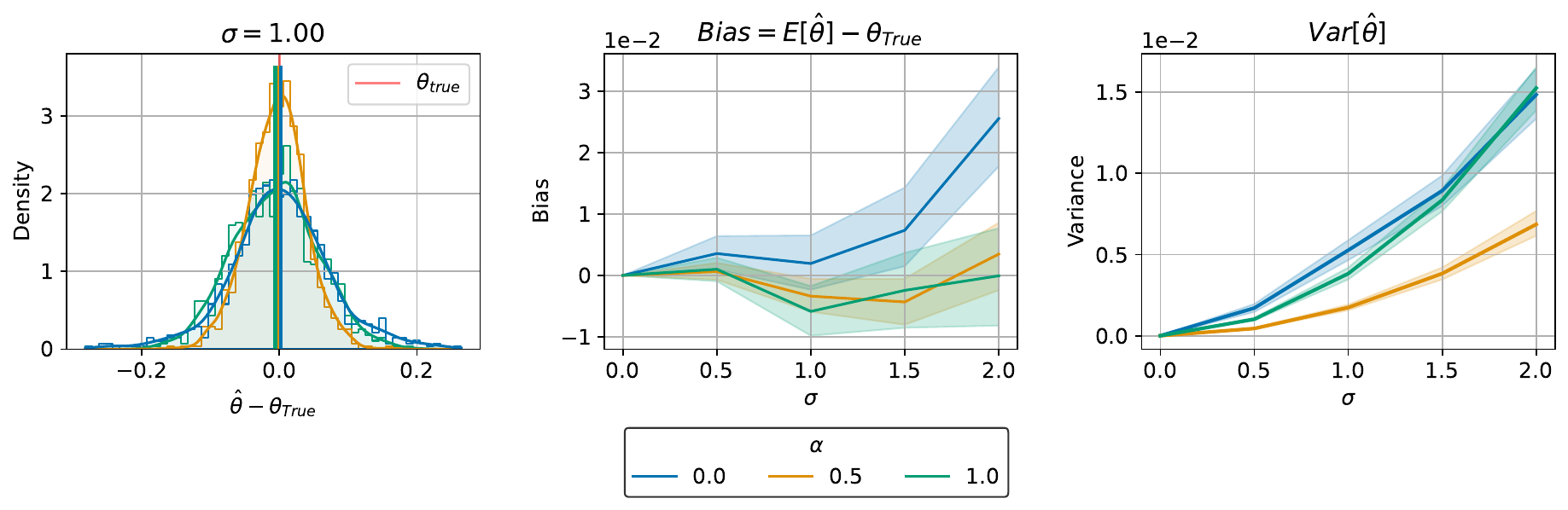}
\vskip -0.1in
\caption{The left panel shows the density distribution of $\hat{\theta} - \theta_{true}$ for a fixed $\sigma$ of 1.0. The middle panel delineates the bias of the estimator, defined as $E[\hat{\theta}] - \theta_{true}$, across varying levels of $\sigma$, and weights $\alpha \in \{0, 0.5, 1\}$ indicated by color. The right panel presents the variance of the estimator, $Var[\hat{\theta}]$, as a function of $\sigma$ for the same set of $\alpha$ values. The shaded regions represent the 95\% bootstrap confidence intervals across ten $\theta_{true}$ values and 100 Monte Carlo simulations.}

\vskip -0.1in
\label{fig:bias_variance}
\end{figure*}

To understand the observed results, we compute the closed-form solutions of the multi-step loss in the case of $\alpha=0$ and $\alpha=1$:

\begin{proposition}
\label{prop:alpha_one}
($\alpha = 1$). Given a transition $(s_t \neq 0, o_{t+1})$ from the linear system and a linear model with parameter $\theta$, the minimizer of the $\alpha=1$ multi-step loss can be computed as:
\[
    \hat{\theta}_1 = \frac{o_{t+1}}{s_t} = \theta_{true} + \frac{\epsilon_{t+1}}{s_t}
\]
\end{proposition}

\begin{proposition}
\label{prop:alpha_zero}
($\alpha = 0$). Given a transition $(s_t \neq 0, o_{t+1}, o_{t+2})$ from the linear system, a linear model with parameter $\theta$, the sign of the true parameter (for instance $\theta_{true} > 0$), and assuming its existence ($ \frac{o_{t+2}}{s_t} > 0$), the minimizer of the $\alpha=0$ multi-step MSE loss can be computed as:
\[
    \hat{\theta}_0 = \sqrt{\frac{o_{t+2}}{s_t}} = \sqrt{\theta_{true}^2 + \frac{\epsilon_{t+2}}{s_t}}
\]
\end{proposition}

\begin{remark}
For ease of notation in \cref{prop:alpha_one} and \cref{prop:alpha_zero}, we compute the solutions given only one transition $(s_t, o_{t+1}, o_{t+2})$. 
In practice, one minimizes the empirical risk based on a training dataset of size $N$: $\cD = \{ (s_{i,t}, o_{i,t+1}, o_{i,t+2}) \}_{i=1:N}$, in which case the closed-form solutions become:
\[
    \begin{cases}
        \hat{\theta}_1 = \theta_{true} + \frac{\sum_{i=1}^N\epsilon_{i,t+1}}{\sum_{i=1}^N s_{i,t}} \\
        \hat{\theta}_0 = \sqrt{\theta_{true}^2 + \frac{\sum_{i=1}^N\epsilon_{i, t+2}s_{i,t}}{\sum_{i=1}^N s_{i,t}^2}} 
    \end{cases}
\]
\end{remark}

On the one hand, we observe that while $\hat{\theta}_1$ is an unbiased estimator of $\theta_{true}$ ($E_{\epsilon_{t+1} \sim \cN(0, \sigma^2)}[\hat{\theta}_1] = \theta_{true}$), its variance grows linearly with the noise scale: $\Var_{\epsilon_{t+1} \sim \cN(0, \sigma^2)}[\hat{\theta}_1] = \frac{\sigma^2}{s_t^2}$. 
On the other hand, $\hat{\theta}_0$ is a potentially biased estimator, but with a smaller variance if $\theta_{true} \gg 1$ ($\frac{\epsilon_{t+2}}{4 \theta_{true}^2 s_t^2} \approx 0$). In this case we can use a first-order Taylor expansion to approximate $\Var[\hat{\theta}_0] = \Var_{\epsilon_{t+2} \sim \cN(0, \sigma^2)}[\hat{\theta}_0]$:

\vskip -0.2in
\begin{align*}
    \Var[\hat{\theta}_0] &= \Var\big[\sqrt{\theta_{true}^2 + \frac{\epsilon_{t+2}}{s_t}}\big] \\
     &\approx \theta_{true}^2 \Var\big[1 + \frac{\epsilon_{t+2}}{2\theta_{true}^2 s_t}\big] = \frac{\sigma^2}{4 \theta_{true}^2 s_t^2} \\
     &\leq \Var[\hat{\theta}_1]
\end{align*}


For intermediate models ($\alpha \in (0,1)$), and in general when the conditions of the last result do not necessarily hold, we use Monte Carlo simulations to compare the variance of $\hat{\theta}_{\alpha \in \{0, 0.5, 1\}}$. 
\cref{fig:bias_variance} shows the variance reduction brought by the multi-step loss when $\alpha = 0.5$. 
It is also noticeable that, up to the noise scales considered in this experiment, $\alpha=0$ generates a large bias (which matches the theoretical insights), while no significant bias is observed for the estimator with $\alpha=0.5$. We conclude that in the case of a noisy linear system, the multi-step MSE loss minimizer with $\alpha=0.5$ is a statistical estimator that (empirically) has a smaller variance and a comparable bias to the one-step loss minimizer. It is worth noticing that when the noise is non-zero the best solution for the one-step MSE is obtained for $\alpha \in (0, 1)$ and not $\alpha=1$ (which corresponds to optimizing exactly the one-step MSE).

In this linear case, we had access to closed-form solutions. However, \cref{fig:l_and_dl} shows that choosing the multi-step MSE loss as an optimization objective introduces additional critical points where a gradient-based optimization algorithm might get stuck. We now empirically study the optimization process in the case of a two-parameter neural network.

\subsection{Two-parameter non-linear system}
\label{subsec:Bi}


As an attempt to get closer to a realistic MBRL setup where neural networks are used for dynamics learning, we study a non-linear dynamical system using a two-parameters neural network model:

\begin{definition}
\label{def:bi}
(Two-parameter non-linear system with additive Gaussian noise). For an initial state $s_0 \in \R$ and unknown parameters $\btheta^{true} = (\theta_1^{true},\theta_2^{true}) \in \R^2$ we define the transition function and observations as
\begin{align*}
s_{t+1} &= \theta_1^{true} \cdot sigmoid(\theta_2^{true} \cdot s_t)  \\ 
o_{t+1} &= s_{t+1} + \epsilon_{t+1} \text{ with } \epsilon_{t+1} \rLeadsto \cN(0, \sigma^2) \text{ and } \sigma \in \R
\end{align*}
\end{definition}

\begin{definition}
\label{def:model_bi}
(Two-parameters neural network model). A single-neuron two-layer (without bias) neural network. We denote its parameters $\btheta = (\theta_1,\theta_2) \in \R^2$:
\begin{align*} 
\hat{s}_{t+1} &= \hat{p}(s_t) = \theta_1 \cdot sigmoid(\theta_2 \cdot s_t)
\end{align*}
\end{definition}

To get a broader idea about the optimization challenges in this setup, we show the loss landscape for different values of $\alpha$ and noise scales $\sigma$ in \cref{appendix:bi_loss_landscape}. 
Regarding the gradient-based optimization, we compare the solution reached after training in terms of the one-step and two-step \emph{validation} MSE losses (the details of the experiment are given in \cref{appendix:bi_exp}). 

\begin{figure}[ht]
\vskip -0.1in
\begin{center}
\centerline{\includegraphics[width=0.8\columnwidth]{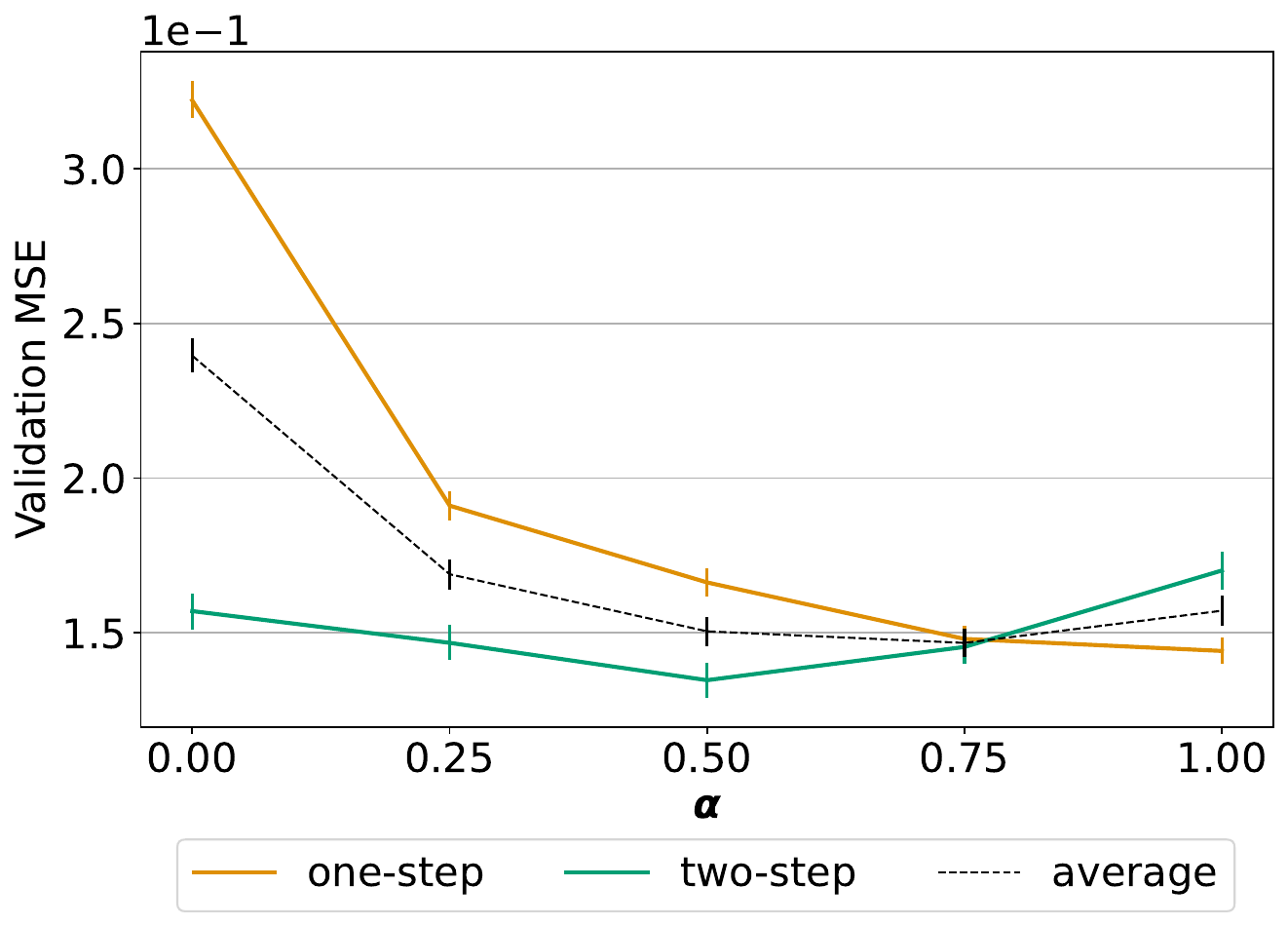}}
\caption{The validation one-step MSE $L_1$ (in yellow), the validation two-step MSE $L_0$ (in green) and the average of these two MSEs (dashed black line) for different values of $\alpha$. The error bars represent the 95\% bootstrap confidence intervals across 2 optimizers, 3 initialization distributions, 10 initial points, 3 noise levels, and 10 Monte Carlo simulations.}
\label{fig:bi_nn_ablation}
\end{center}
\vskip -0.2in
\end{figure}

\cref{fig:bi_nn_ablation} shows that even in the presence of noise, compared to the linear case, the best value of $\alpha$ for the one-step MSE is $\alpha=1$. This can be explained by the use of lower levels of noise in the simulations than the linear case.
The intermediate models obtained for $\alpha \in \{0.25, 0.5, 0.75\}$ represent a trade-off between the one-step and two-step MSEs. The average of the two MSEs, which we use in the experiment section to assess the overall quality of our models over a range of horizons, achieves its minimum for $\alpha=0.75$. 
Notice that the validation losses are only used for evaluation, and do not match the training loss which depends on $\alpha$.



\section{Experiments \& results}
\label{sec:Experiments}

In this section, we evaluate the performance of models trained with the multi-step loss $L_{\balpha}$ for different values of the horizon $h$ and different noise levels of the dynamics. We first describe the multi-step loss hyperparameters: the horizon $h$ and the weights $\balpha$. For the evaluation of the models we consider both a static and a dynamic setup. The \emph{static evaluation} denotes the evaluation of dynamics models in terms of predictive error in held-out test datasets. For the \emph{dynamic evaluation} we will consider the offline MBRL setting \citep{Levine2020} where the goal is to learn a policy from a given dataset without interacting further with the environment. The evaluation is done on three classical RL environments (\emph{Cartpole swing-up}, \emph{Swimmer} and \emph{Halfcheetah}) and various datasets (eight in total) collected with different behavior policies (\emph{random}, \emph{medium}, \emph{full\_replay} and \emph{mixed\_replay}) on these environments. The details of these tasks are provided in \cref{appendix:setup}. For all these tasks, we use the same neural network model for $\hat p_\theta$. Implementation details for the model are given in \cref{appendix:imp}.


\subsection{Hyperparameters of the multi-step loss}
\label{subsec:hypers}

The multi-step loss depends on the following hyperparameters: the loss horizon $h$ and the corresponding weights $\{\alpha_i\}_{i=1}^h$.

\subsubsection{The horizon $h$}
\label{subsubsec:h}

The horizon of the loss is the maximal prediction horizon considered in the loss. Typically, in this paper we consider $h \in \{2,3,4,10\}$ for the multi-step loss and $h=1$ for the single-step baseline.

\subsubsection{The weights \(\alpha\)}
\label{subsubsec:alpha}

For a horizon $h$, the weights $\balpha$ live in the $(h-1)$-dimensional probability simplex, which means that the search space is $(0,1)^{h-1}$. As the MSE has been shown to grow exponentially with the horizon when applying one-step models recursively (Theorem 1 of \citet{Venkatraman_2015}) we choose an exponentially parameterized profile for the weights. The objective is to give each MSE term the same importance when we optimize the MSE loss. Given a weighted multi-step loss $L_\balpha$ with horizon $h$, weights $\balpha$, and $\beta > 0$, the exponentially parameterized weights are defined as:
\[
\alpha_i = \underbrace{(\frac{1-\beta}{1-\beta^{h+1}})}_{\text{normalization constant}} \cdot \beta^i \hspace{0.3cm} \text{for } i \in \{1,\hdots,h\} .
\]

Applying this exponential parametrization of the weights, we further reduce the search space dimension. In our experiments, we still consider values greater than 1 in the grid, values corresponding to an increase of the weights over the horizon. This approach is used to analyze whether certain settings would benefit from focusing on large horizons.

\subsection{Static evaluation with the R2 metric}
\label{subsec:static}

The commonly used metrics for the static evaluation are the standard mean squared error (MSE) or the explained variance (R2) which we prefer over the MSE because it is normalized and can be aggregated over multiple dimensions. In our attempt to reduce compounding errors in MBRL, we are especially interested in the long-term predictive error of models. For each horizon $h$, the error is computed by considering all the sub-trajectories of size $h$ from the test dataset. The predictions are computed by calling the model recursively $h$ times (using the ground truth actions of the sub-trajectories) and the average R2 score at horizon $h$, $\text{R2}(h)$, averaged over the sub-trajectories is reported: given a dataset $\cD=\{(s_{i,t},a_{i,t},s_{i,t+1},\hdots)\}_{i=1,\hdots,N}$, a predictive model $\hat{p}_\theta$, for $h=1,\hdots,H$:
\begin{equation*}
    \text{R2}(h) = \frac{1}{d_s} \sum_{j=1}^{d_s} 1 - \frac{\frac{1}{N} \sum_{i=1}^{N} \left(s^j_{i,t+h} - \hat{p}_\theta^h(s_{i,t},\ba_{i, t:t+h})^j\right)^2}{\frac{1}{N} \sum_{i=1}^{N} \left(s^j_{i,t+h} - \Bar{s}^j_{t+h} \right)^2}.
\end{equation*}
We also report the average $\text{R2}$ score, $\overline{R2}(H)$, over all prediction horizons from $1$ to $H$.
\begin{equation*}
    \overline{R2}(H) = \frac{1}{H} \sum_{h=1}^H \text{R2}(h).
\end{equation*}


For the weights of the loss we perform a grid search over $\beta$ values, selecting the value giving the best $\overline{R2}(H)$ averaged over 3 cross-validation folds.

\begin{figure*}[htp]
\centering
\includegraphics[width=.99\textwidth]{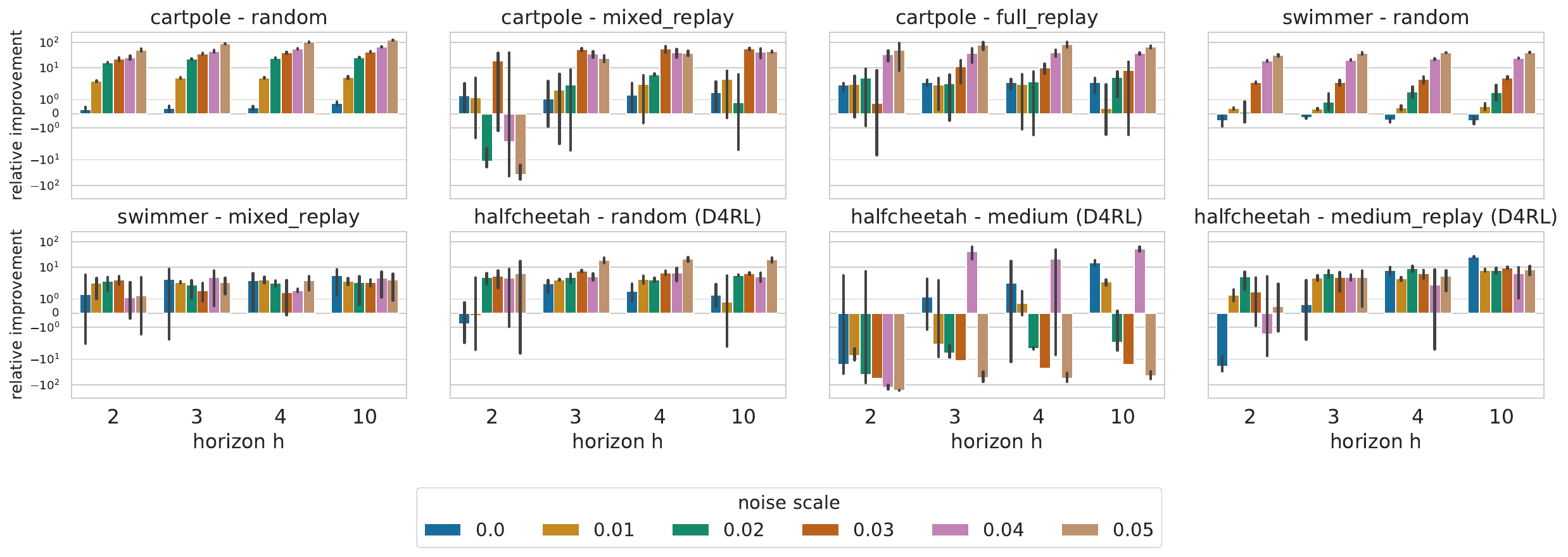}
\vskip -0.1in
\caption{The series of bar plots display the relative improvement with respect to the $h=1$ baseline, in the test $\overline{R2}(50)$ metric for various environments, and datasets. Performance is evaluated over loss horizons $h$ with the relative improvement measured on a logarithmic scale. The relative noise scale, ranging from 0.0 to 0.05, is color-coded and represents a ratio of the range of the state space, for each dataset. The error bars indicate the 95\% confidence intervals ($\text{mean} \pm 1.96 \cdot \text{standard error}$) obtained with the three cross-validation folds.}
\vskip -0.15in
\label{fig:relative_all}
\end{figure*}

\cref{fig:relative_all} shows the relative improvement in percentage over the one-step model of the $\overline{R2}(50)$ metric for the different datasets (the absolute values of the metric and the full $R2(h)$ curves are shown in \cref{appendix:static_section}). For most of the datasets (all the \emph{Cartpole} and \emph{Swimmer} datasets, and \emph{Halfcheetah} \emph{random} and \emph{medium\_replay}) the benefit of using the multi-step loss when there is noise is clear and for most of them (all the \emph{Cartpole} datasets, \emph{Swimmer random} and \emph{Halfcheetah random}) the larger the noise the higher the benefit. The impact of the horizon $h$ of the loss is less clear although for some datasets increasing the horizon $h$ as the noise increases also helps. This result is more mitigated on \emph{Halfcheetah medium} which we suggest is due to the optimization process converging to sub-optimal solutions. We also note that even if the multi-step loss is using $h=10$, because of the weights, the multi-step loss could finally be close to the multi-step losses obtained with $h=2$, 3 or 4, which would mean that a smaller horizon is sufficient. This is an advantage of our flexible multi-step loss that can adapt its horizon thanks to the selection of the best weights

We now study the best weights obtained for each multi-step loss ($h=2$, 3, 4, and 10) and each noise scale. First, we observe from \cref{table:beta} that compared to what is usually done in the literature, the best weights are not uniform. Second, we also notice an increase of the parameter $\beta$ with the noise scale. This finding supports the idea that multi-step models are increasingly needed when incorporating information from the future is crucial to achieve better performance. We also discuss the effective horizon obtained with the weights in \cref{appendix:effective_horizon}.

\begin{table}[!htp]
\small
\caption{Best $\beta$ values found with a grid search for each horizon and each noise scale. The values are averaged over the eight datasets.}
\label{table:beta}
\centering
\vspace{0.5em}
\begin{tabular}{l|llll}
\toprule
& \multicolumn{4}{c}{horizon $h$} \\ \midrule
 & 2 & 3 & 4 & 10 \\ \midrule
$\sigma$ (\%) & \multicolumn{4}{c}{$\beta$} \\ \hline
0 & 0.81 & 0.45 & 0.41 & 0.46 \\
1 & 0.56 & 0.76 & 0.51 & 0.47 \\
2 & 0.86 & 0.72 & 0.78 & 0.54 \\
3 & 0.67 & 1.21 & 0.81 & 0.50 \\
4 & 0.74 & 1.01 & 0.74 & 0.55 \\
5 & 1.33 & 0.83 & 0.97 & 0.51 \\ \bottomrule
\end{tabular}
\end{table}

\subsection{Dynamic evaluation: offline MBRL}
\label{subsec:mbrl}

We consider the offline setting where given a set of $N$ trajectories $\mathcal{D}=\{(s_t^i,a_t^i,s_{t+1}^i,\hdots)\}_{i=1}^N$, the goal is to learn a policy maximizing the return in a single shot, without further interacting with the environment.

We consider a \emph{Dyna}-style agent that learns a parametric model of the policy based on data generated from the learned dynamics model $\hat{p}_\theta$. Specifically, we train a Soft Actor-Critic (SAC) \citep{Haarnoja2018} with short rollouts on the model a la MBPO \cite{Janner2019}, a popular MBRL algorithm. We then rely on model predictive control (MPC) at decision-time where the action search is guided by the SAC policy. The details of this agent are given in \cref{appendix:agents}.

The experiments are run on the \emph{Cartpole mixed\_replay} dataset without noise (noise scale of $1\%$). The reason is that the range of episode returns spanned by this dataset makes us hope that it is possible to learn a model that is good enough to learn a successful policy (for which episode returns are higher than 800). The distribution of the returns on the random and full replay datasets makes it more challenging to learn a successful policy. The goal here is not to study a new offline MBRL algorithm, avoiding unknown regions of the state-action space, but studying the improvement that can be obtained with the multi-step loss when varying the horizon $h$. Finally, in order to isolate the effect of dynamics learning, we assume that the reward function is a known deterministic function of the observations.

\begin{figure}[b]
\vskip -0.1in
\begin{center}
\centerline{\includegraphics[width=0.9\columnwidth]{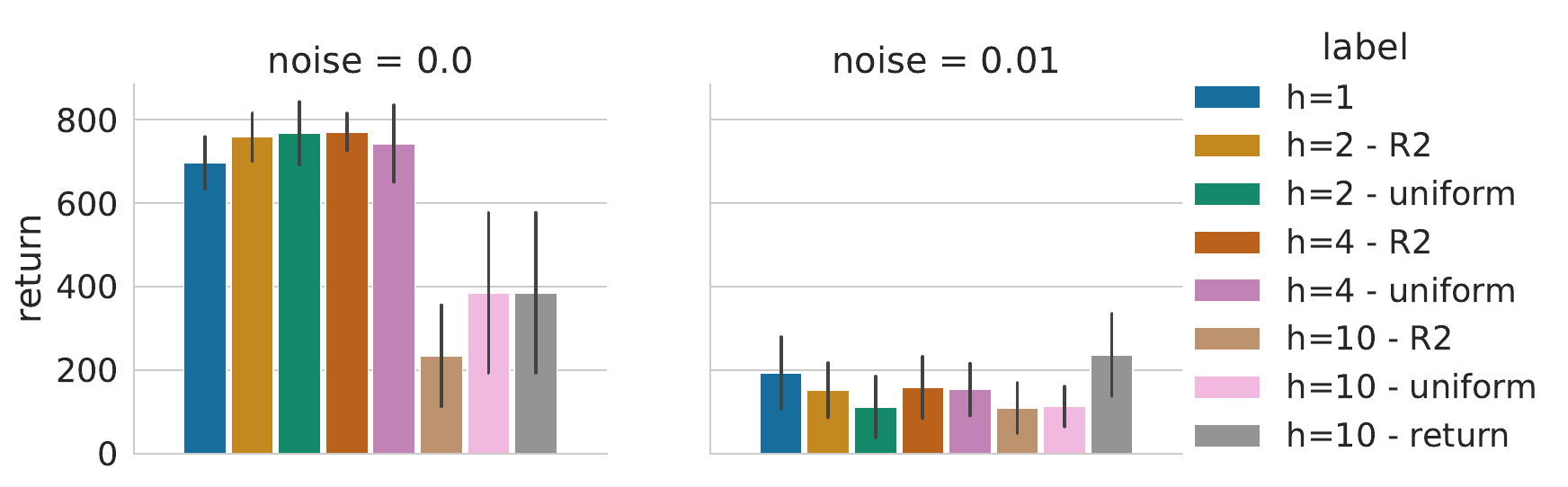}}
\caption{Returns of the agents trained on multi-step models in the \emph{Cartpole mixed\_replay} task. The error bars indicate the 95\% confidence intervals ($\text{mean} \pm 1.96 \cdot \text{standard error}$).}
\label{fig:abs_return}
\end{center}
\vskip -0.2in
\end{figure}

As it is acknowledged that static evaluation metrics may not always align with the final return of agents (see for instance \citep{Lambert2021}) we also include the performance of the models trained on the 10-step loss with weights tuned to maximize the return, using the same grid search as for the R2.

\paragraph{Without noise.}
\cref{fig:abs_return} shows the return obtained by the different models on the \emph{Cartpole mixed\_replay} task. Against a strong baseline ($h=1$) with a near 700 return, it is observed that the multi-step models exhibit marginally superior performance, especially for $h$ values of 2 and 4. However, this trend is not maintained for $h=10$, indicating that larger horizons may not be beneficial in this context. Additionally, the study compares these outcomes with those obtained using uniform weights, a common approach in most related research. As depicted in \cref{fig:abs_return}, the returns for uniform weight variants at $h$ values of 2 and 4 are comparable to those of the models selected based on the $\overline{R2}(50)$ metric. While uniform weights help the $h=10$ model reach a better return, it remains sub-optimal with respect to the baseline. These findings confirm a common result in the literature \citep{Lutter2021, Xu2018}, stating that the multi-step loss is only useful for small values of $h$. 

\paragraph{With noise.} In the presence of noise, we observe that the performance of all models, including the baseline, significantly decreases. Notably, the optimal multi-step models do not demonstrate any improvement in this setting, a finding that also holds true for the variant with uniform weights. We found that $\beta=0.3$ maximizes the return for the 10-step loss and leads to a marginal improvement over the baseline model. This finding challenges the common practice of using uniform weights in related research, suggesting that a more tailored approach is necessary for each specific task. We believe that a better hyperparameter search for the multi-step loss weights could lead to better performance of the corresponding multi-step models.


\section{Conclusion}
\label{sec:Conclusion}

In this paper, we introduce a weighted multi-step loss that leads to models exhibiting significant improvements in the average R2 score over future horizons in various datasets derived from popular RL environments with noisy dynamics.
These outcomes align with our analysis of two tractable scenarios: models optimized for the multi-step loss outperform those focused on minimizing the one-step loss in the presence of noise. The insights from dynamic evaluation present a more complex picture. Specifically, we found that in absence of noise, the multi-step loss slightly improved the return of an already strong one-step baseline. However, a noticeable improvement in the noisy scenario would require an extensive search for optimal hyperparameters, including the weights and, implicitly, the prediction horizon.




\newpage


\bibliography{main}
\bibliographystyle{main}

\newpage
\appendix
\onecolumn

\addcontentsline{toc}{section}{Appendix} 
\part{Appendix} 
\parttoc 

\section{The evaluation setup}
\label{appendix:setup}

\subsection{Environments}
\label{appendix:envs}

\begin{wrapfigure}{r}{.5\textwidth}
    \begin{minipage}{\linewidth}
    \centering
    \subfigure[Cartpole]{\includegraphics[width=.3\textwidth]{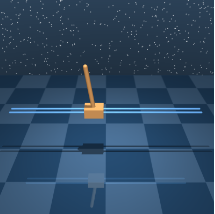}}
    \subfigure[Swimmer]{\includegraphics[width=.3\textwidth]{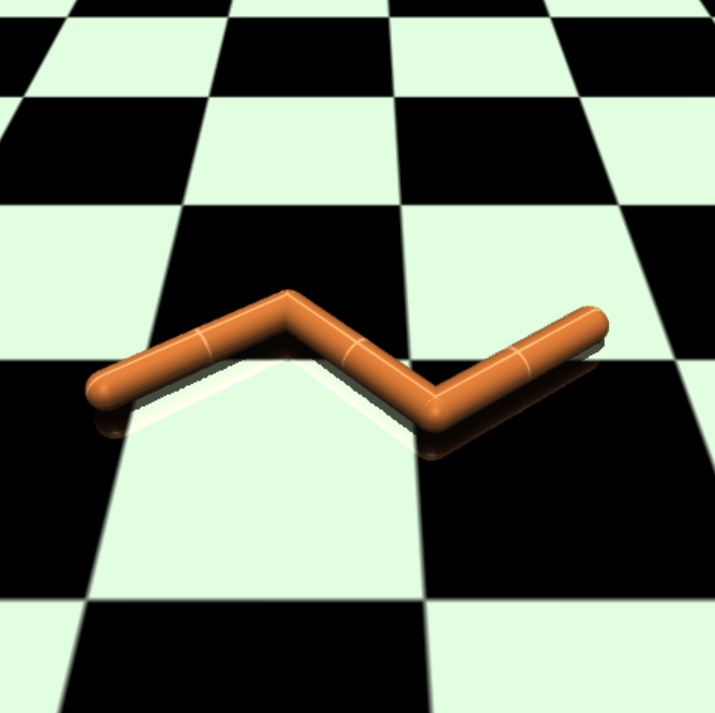}}
    \subfigure[Halfcheetah]{\includegraphics[width=.3\textwidth]{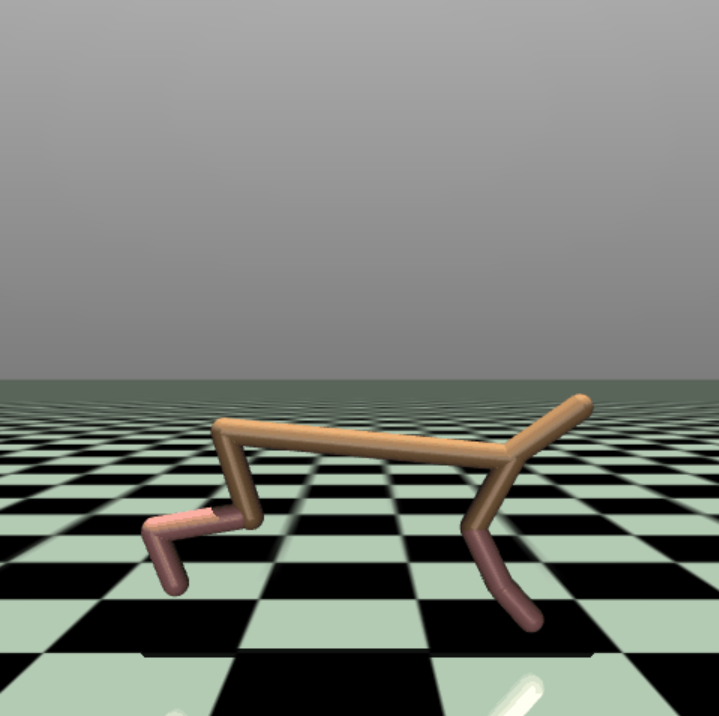}}
\end{minipage}
\vskip -0.05in
\caption{The environments: Cartpole swing-up, Swimmer and Halfcheetah.}
\vskip -0.1in
\label{fig:envs}
\end{wrapfigure} 

In the present study, we examine three distinct environments within the scope of continuous control reinforcement learning, as delineated in \cref{fig:envs}, each exhibiting varying degrees of complexity. The complexity of a given environment is primarily determined by the dimension of the state space $d_s$, and the dimension of the action space $d_a$. Notably, \emph{Cartpole swing-up} is a classic problem in the field where the task is to swing up a pole starting downwards, and balance it upright. The other considered tasks are \emph{Swimmer} and \emph{Halfcheetah}, which are two locomotion tasks, aiming at maximizing the velocity of a virtual robot along a given axis. \emph{Swimmer} incorporates fluid dynamics with the goal of learning an agent that controls a multi-jointed snake moving through water. On the other hand, \emph{Halfcheetah} simulates a two-leg Cheetah with the goal of making it run as fast as possible. For the latter environments, we use the implementation of OpenAI Gym \citep{Brockman2016}, and the implementation of Deepmind Control \citep{Tassa_2018} for \emph{Cartpole}. Both libraries are based on the Mujoco physics simulator \citep{Todorov2012}. A detailed description of these environments is provided in \cref{table:envs}.

\begin{table}[!ht]
  \scriptsize
  \caption{The environments characteristics. $d_s$: the dimension of the state space, $d_a$: the dimension of the action space, $x_t$: position along the $x$-axis, $\Dot{x}_t$: velocity along the $x$-axis, $\|a_t\|_2^2$: the action $a_t$ magnitude, $\theta$: the angle of the pole (only for Cartpole).}
  \label{table:envs}
  \centering
  \begin{tabular}{ccccc}
    \toprule
    environment 
    & $d_s$
    & $d_a$
    & task horizon
    & reward function
    \\
    \cmidrule(r){1-5}
    Cartpole swing-up
    & 5
    & 1 
    & 1000
    & $\frac{1 + \cos \theta_t}{2} 
\times \frac{1 + e^{-0.25\log (10)x_t^2}}{2}
\times \left(1 - \frac{a_t^2}{5}\right)
\times \frac{1 + e^{-0.04\log (10){\dot x_t}^2}}{2}$
    \\
    Swimmer
    & 8
    & 2 
    & 1000 
    & $\Dot{x}_t - 0.0001 \times \|a_t\|_2^2$
    \\
    Halfcheetah
    & 17
    & 6 
    & 1000 
    & $\Dot{x}_t - 0.1 \times \|a_t\|_2^2$
    \\
    \bottomrule
  \end{tabular}
\vspace{-2mm}
\end{table}

\subsection{Datasets}
\label{appendix:datasets}

In this section, we introduce the different datasets that are used to evaluate the multi-step models. These datasets are collected using some \textit{behavior policies} that are unknown to the models. 

\cref{table:datasets} illustrates the features of datasets across the three environments: Cartpole Swing-up, Swimmer, and Halfcheetah. These environments vary in dataset size and behavioral policies. In the Cartpole Swing-up setting, each of the three datasets (\emph{random}, \emph{mixed\_replay}, and \emph{full\_replay}) includes 50 episodes, which are split into training, validation, and testing subsets. The  and \emph{full\_replay} depict complete learning trajectories of an unstable model and a state-of-the-art (sota) model-based Soft Actor-Critic (SAC) respectively \citep{Janner2019}, and integrated with shooting-based planning \cref{appendix:agents}. For the Swimmer environment, both the random and \emph{full\_replay} datasets consist of 50 episodes each. The random dataset is derived from a random policy, whereas the \emph{full\_replay} dataset is generated using a model-based SAC with planning. For both the Cartpole Swing-up and Swimmer environments, the datasets were self-collected due to the absence of a unified benchmark that includes datasets from both tasks.

\begin{table}[!ht]
  \scriptsize
  \caption{The datasets characteristics. \emph{mf}: model-free, \emph{mb}: model-based, random $\rightarrow \pi$: all episodes collected to learn the policy $\pi$. The datasets size is given in episodes (of 1000 steps each).}
  \label{table:datasets}
  \centering
  \begin{tabular}{cccc}
    \toprule
    environment 
    & dataset
    & size (train/valid/test)
    & behavior policy
    \\
    \midrule
    \multirow{3}{*}{Cartpole swing-up} 
    & random
    & 50 (36/4/10) 
    & random policy
    \\
    & mixed\_replay
    & 50 (36/4/10)
    & random $\rightarrow$ unstable \emph{mb} SAC + planning
    \\
    & full\_replay
    & 50 (36/4/10)
    & random $\rightarrow$ \emph{mb} SAC + planning
    \\
    \midrule
    \multirow{2}{*}{Swimmer} 
    & random
    & 50 (36/4/10) 
    & random policy
    \\
    & mixed\_replay
    & 50 (36/4/10)
    & random $\rightarrow$ unstable \emph{mb} SAC + planning
    \\
    \midrule
    \multirow{3}{*}{Halfcheetah} 
    & random (\emph{D4RL})
    & 100 (76/4/20) 
    & random policy
    \\
    & medium (\emph{D4RL})
    & 100 (76/4/20) 
    & \emph{mf} sac at half convergence
    \\
    & medium\_replay (\emph{D4RL})
    & 200 (156/4/40) 
    & random $\rightarrow$ \emph{mf} sac at half convergence
    \\
    \bottomrule
  \end{tabular}
\vspace{-2mm}
\end{table}

\begin{figure}[ht]
\begin{center}
   \includegraphics[width=.95\linewidth]{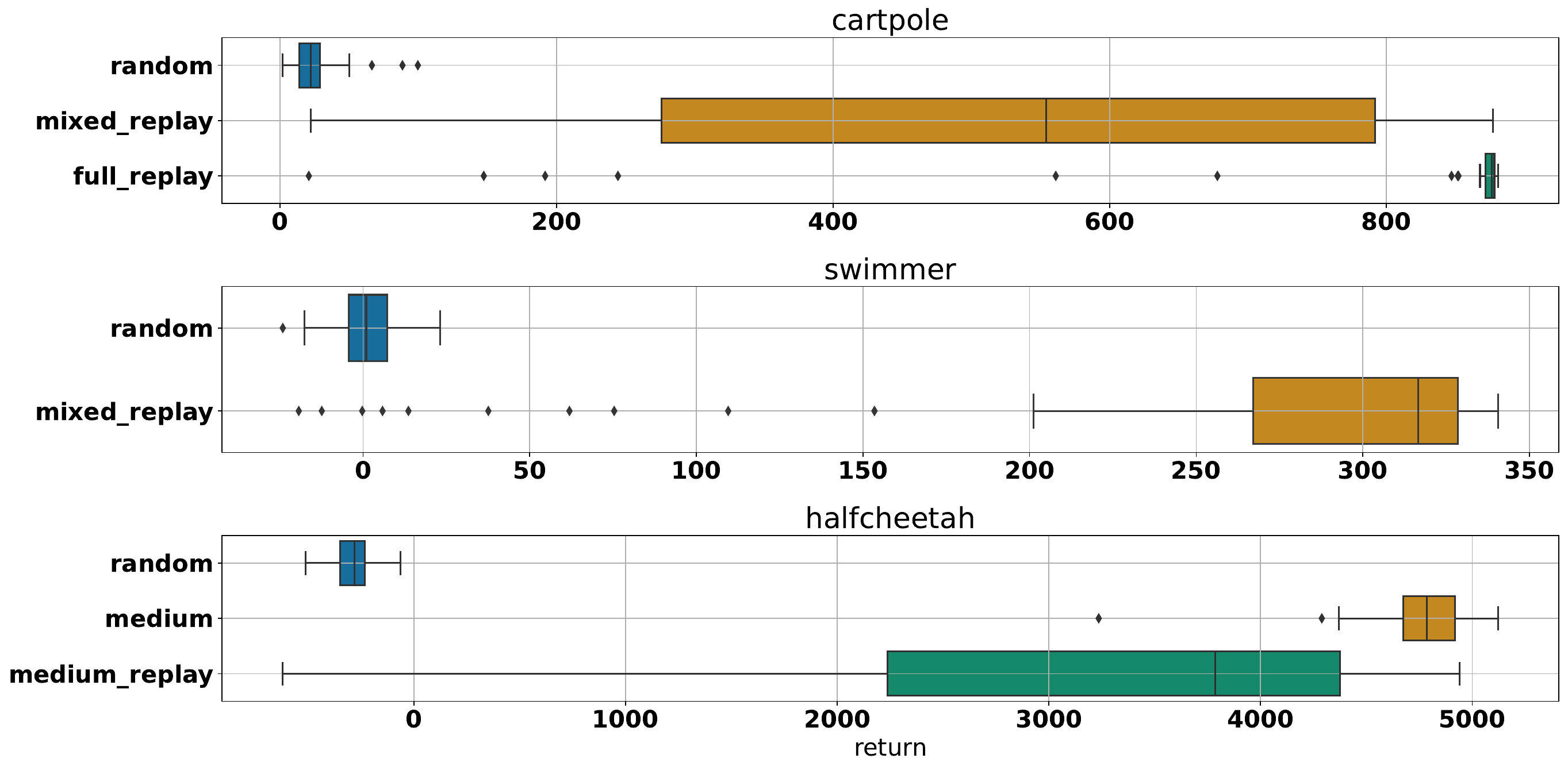}
\end{center}
\caption{A comparison of the distribution of returns across the considered datasets.}
\label{fig:DataReturn}
\end{figure}

To enhance our understanding of the differences among these datasets, we present the distribution of returns for each dataset in \cref{fig:DataReturn}. It is important to note that the variance in returns within a dataset serves as an indicator of the extent of the state space covered by that dataset. Specifically, datasets collected using a fixed policy exhibit a notably narrow distribution, predominantly concentrated around their mean values, as exemplified by the Halfcheetah \emph{random} and \emph{medium} datasets. This characteristic of the datasets significantly influences the out-of-distribution generalization error in offline MBRL, which represents a major challenge in this context.

\subsection{Agent: SAC + planning}
\label{appendix:agents}

Soft Actor-Critic (SAC) \citep{Haarnoja2018} is an off-policy algorithm that incorporates the maximum entropy framework, which encourages exploration by seeking to maximize the entropy of the policy in addition to the expected return. SAC uses a deep neural network to approximate the policy (actor) and the value functions (critics), employing two Q-value functions to mitigate positive bias in the policy improvement step typical of off-policy algorithms. This approach helps in learning more stable and effective policies for complex environments, making SAC particularly suitable for tasks with high-dimensional, continuous action spaces.

In addition to \emph{Dyna}-style training of the SAC agent on the learned model with short rollouts a la MBPO \citep{Janner2019}, we use Model Predictive Control (MPC). MPC is the process of using the model recursively to plan and select the action sequence that maximizes the expected cumulative
reward over a planning horizon $H$. The set of $K$ (population size) action sequences $ \{(a^k_{t:t+H})_{k \in \{1,\dots,K\}}\}$ is usually generated by an evolutionary algorithm, e.g. Cross Entropy Method (CEM) \citep{Chua2018}. In this study, the pre-trained SAC guides the MPC process by generating candidate action sequences from the learned stochastic policy. 

\section{Implementation details}
\label{appendix:imp}

For all the models, we use a neural network composed of a common number of hidden layers and two output heads (with \emph{Tanh} activation functions) for the mean and standard deviation of the learned probabilistic dynamics (The standard deviation is fixed when we want to use the MSE loss). We use batch normalization \citep{Ioffe2015}, Dropout layers \citep{Srivastava2014} ($p=10\%$), and set the learning rate of the Adam optimizer \citep{Diederik2015} to $0.001$, the batch size to $64$, the number of common layers to $2$, and the number of hidden units to $256$ based on a hyperparameter search executed using the RAMP framework \citep{Kegl2018}. The evaluation metric of the hyperparameter optimization is the aggregated one-step validation R2 score across all the offline datasets. The neural networks are trained to predict the difference between the next state and the current state $\Delta_{t+1} = s_{t+1} - s_t$. More precisely, the baseline consists in the single-step model trained to predict the difference $\Delta_{t+1}$ using the one-step MSE. The other multi-step models, take their own predictions as input to predict the difference $\Delta_{t+h} = s_{t+h} - \hat{s}_{t+h-1}$ at horizon $h$

For the offline RL experiments, we use SAC agents from the StableBaselines3 open-source library \citep{Raffin2021} while keeping its default hyperparameters. In the offline setting, we train the SAC agents for $500,000$ steps on a fixed model by generating short rollouts of length 100 from states of the the dataset selected uniformly at random. At evaluation time, the MPC planning is done by sampling $500$ action sequences from the SAC policy, and rolling out short rollouts of horizon $20$ for return computation. This return is then bootstrapped with the value function learned by SAC.  

\section{Probabilistic interpretation}
\label{appendix:Probabilistic}

So far we have only considered deterministic models that directly learn to predict the next observation. However, it is very common in the literature to use \emph{probabilistic} models that learn to predict the parameters of a \emph{dist   ribution} over the next observations \citep{Chua2018, Janner2019, Nagabandi2018, Kegl2021}, rather than a point estimate. Probabilistic models are useful because they enable uncertainty estimation, which can represent finite-sample fitting errors (epistemic uncertainty) and/or the intrinsic uncertainty in the environment dynamics (aleatory uncertainty). 

In the probabilistic case, we represent $\hat{p}_\theta$ as a Gaussian over the next state
\[
\hat{p}_\theta\big(s_{t+1} | s_t, a_t \big) = \cN\big(s_{t+1}|\hat{\mu}_\theta(s_t,a_t),\hat{\sigma}_\theta(s_t,a_t)\big),
\]
where the parameters $(\hat{\mu}_\theta, \hat{\sigma}_\theta)$ are the output of the neural network. The loss function to train the probabilistic model on a transition\footnotemark $(s_t,a_t,s_{t+1})$ is the negative log-likelihood $L\big(s_{t+1}, \hat{p}_\theta(\cdot|s_t, a_t)\big) = - \log \hat{p}_\theta(s_{t+1}|s_t, a_t)$. 

Using this formulation, the $j^{th}$-horizon negative log-likelihood loss is
\vskip -0.2in
\begin{align*}
    L\big(s_{t+j}, \hat{p}_\theta^j(s_t, \ba_{t:t+j})\big) & = - \log \hat{p}^j_\theta(s_{t+j}|s_t, \ba_{t:t+j}) \\
     &= - \log \cN\big(s_{t+j}|\hat{\mu}^j_\theta(s_t,\ba_{t:t+j}),\hat{\sigma}^j_\theta(s_t,\ba_{t:t+j})\big).
\end{align*}
\vskip -0.1in

\footnotetext{In this section, we don't make the distinction between states $s$ and observations $o$ as the probabilistic interpretation is independent of the underlying MDP.}

Instead of the direct next prediction, the parameters of the distribution $\big(\hat{\mu}^j_\theta(s_t,\ba_{t:t+j}), \hat{\sigma}^j_\theta(s_t,\ba_{t:t+j})\big)$ represent the output of the model after $j$ recursive calls using the ground truth actions $\ba_{t:t+j}$. From this definition, we can obtain the weighted multi-step MSE loss by solving a Maximum (joint) Likelihood Estimation (MLE) problem: 

\begin{proposition}
\label{prop:likelihood}
(Multi-step MSE loss as MLE). Under the assumption of the conditional independence on $(s_t, \ba_{t:t+h})$, the weighted multi-step MSE loss can be recovered from the negative log-likelihood of the joint distribution of  $(s_{t+1}, \ldots, s_{t+h}) = \bs_{t+1:t+h}$:
\begin{align*}
    L\big(\bs_{t+1:t+h},\hat{p}_\theta^{1:h}(s_t, \ba_{t:t+h})\big) &= - \log \hat{p}^{1:h}_\theta (\bs_{t+1:t+h}|s_t, \ba_{t:t+h}) \\
    &= - \log \Pi_{j=1}^h \hat{p}^j_\theta(s_{t+j}|s_t, \ba_{t:t+j}) \\
     &= \underbrace{\sum_{j=1}^h \log \hat{\sigma}^j_\theta}_{\text{regularization}} + \underbrace{\sum_{j=1}^h \underbrace{\frac{1}{2 (\hat{\sigma}^j_\theta)^2}}_{\alpha_j} (s_{t+j} - \hat{\mu}^j_\theta)^2}_{L_\balpha} + \underbrace{C}_{constant} 
\end{align*}
with $\hat{\mu}^j_\theta = \hat{\mu}^j_\theta(s_t,\ba_{t:t+j})$ and $\hat{\sigma}^j_\theta = \hat{\sigma}^j_\theta(s_t,\ba_{t:t+j})$.
\end{proposition}

\begin{remark}
\cref{prop:likelihood} states the result in the case of uni-dimensional state spaces. While this is considered for simplicity, we can straight-forwardly generalize to the multi-variate case with the inverse of the covariance matrix as weight $ \alpha_j = -\frac{1}{2} \hat{\Sigma}_\theta^{-1}$.
\end{remark}

\paragraph{Can we automatically learn the weights using the MLE formulation in \cref{prop:likelihood}?}

To answer this question, we train a probabilistic model $\hat{p}_\theta$ to minimize the Negative Log-Likelihood (NLL) loss shown at \cref{prop:likelihood}.
When computing the multi-step loss at training time, the predicted states are obtained by sampling from the learned distribution using the reparametrization trick $\hat{s}_{t+j} = \hat{\mu}^j_\theta(s_t,\ba_{t:t+j}) + \hat{\sigma}^j_\theta(s_t,\ba_{t:t+j}) \cdot \xi$ where $\xi \sim \cN(0,\bI)$\footnotemark. We refer to this as \emph{stoch}-astic sampling, as opposed to \emph{det}-erministic sampling that refers to MSE-based models where $ \hat{s}_{t+j} = \hat{\mu}^j_\theta(s_t,\ba_{t:t+j})$.

\footnotetext{
Notice that in this context, $s$ is not necessarily uni-dimensional. In fact, $s \in \R^d$ where $d$ is the dimension of the state space. $\bI$ is thus, the corresponding identity matrix. 
}

\begin{figure}[ht]
\begin{center}
\centerline{\includegraphics[width=\columnwidth]{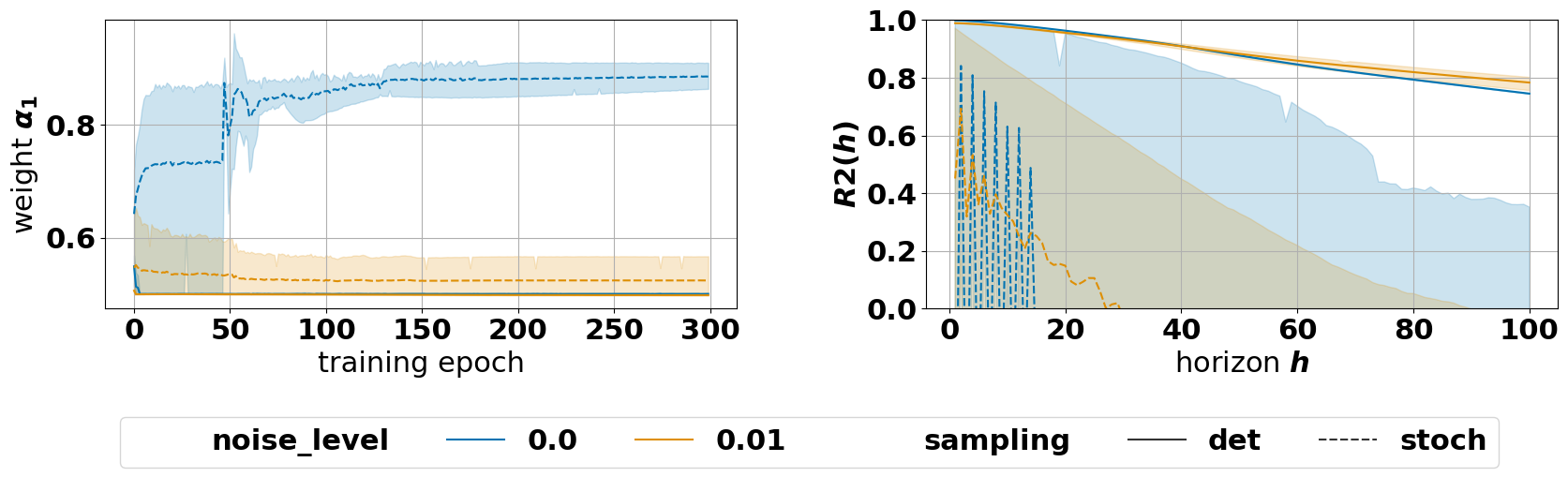}}
\vskip -0.1in
\caption{The left panel shows the normalized learned weights of a stochastic model trained using the $h=2$ multi-step loss. Basically, $\alpha_1 = \frac{\frac{1}{|\cD_{train}|} \sum_{\cD_{train}} \frac{1}{2(\hat{\sigma}^1_\theta)^2}}{\frac{1}{|\cD_{train}|} \sum_{\cD_{train}} \frac{1}{2(\hat{\sigma}^1_\theta)^2} + \frac{1}{|\cD_{train}|} \sum_{\cD_{train}} \frac{1}{2\hat({\sigma}^2_\theta)^2}}$ and $\alpha_2 = 1 - \alpha_1$. The right panel shows the $R2(h)$ curves on a held-out test dataset from the \emph{Cartpole swing-up random} task.}
\label{fig:stoch_vs_det}
\end{center}
\vskip -0.2in
\end{figure}

Interestingly, we find that the \emph{stoch} model automatically learns to put more weight on $h=1$ in absence of noise, while learning to balance the weights at $1\%$ noise (\cref{fig:stoch_vs_det}). In the same dataset (\emph{Cartpole-Random} with respectively $0\%$ and $1\%$ noise scale), we have found for the $h=2$ MSE-based model that the optimal weights were obtained by respectively $\beta=0.1$ ($\alpha_1 = 0.9$ and $\alpha_2 = 0.1$) and $\beta=2.0$ ($\alpha_1 = 0.33$ and $\alpha_2 = 0.66$). This result shows the potential of the MLE formulation to learn the correct weight profile depending on the intrinsic noise on the training data, which is captured through the learned variance.

In practice, the \emph{stoch} model fails to predict accurately at future horizons (\cref{fig:stoch_vs_det}), which is the reason why we don't adopt this formulation in the main paper. 
We suspect that the model is underestimating the variance, leading to inaccurate predictions during inference and resulting in divergent trajectories. This issue stems from inadequate uncertainty propagation during the generation of rollouts from the stochastic model. In fact, accurate uncertainty quantification over future horizons necessitates the generation of multiple trajectories and the subsequent measurement of variance at the $j$-th horizon. We leave the exploration of this promising direction to future work.

\section{Additional experiments \& results}
\label{appendix:experiments}

\subsection{Uni-dimensional linear system}
\label{appendix:uni}

\subsubsection{Comparison with the data augmentation variant}
\label{appendix:uni_data_aug}

To further understand the multi-step loss, we propose to reconduct the bias-variance analysis in \cref{subsec:Uni} with a data augmentation variant that consists in training the $\alpha=1.0$ model (baseline) on an augmented dataset: $\cD_{aug} = \{\cS_0, \cO_1\}$, with the next states being $y = \{\cO_1, \cO_2\}$. 

Indeed, one may argue that the variance reduction gained by using the $\alpha=0.5$ multi-step loss is due to sampling the noise twice (both through $\cO_1$, and $\cO_2$), while the vanilla model only sees one realization of the noise through $\cO_1$.

\begin{figure}[ht]
\begin{center}
   \includegraphics[width=.95\linewidth]{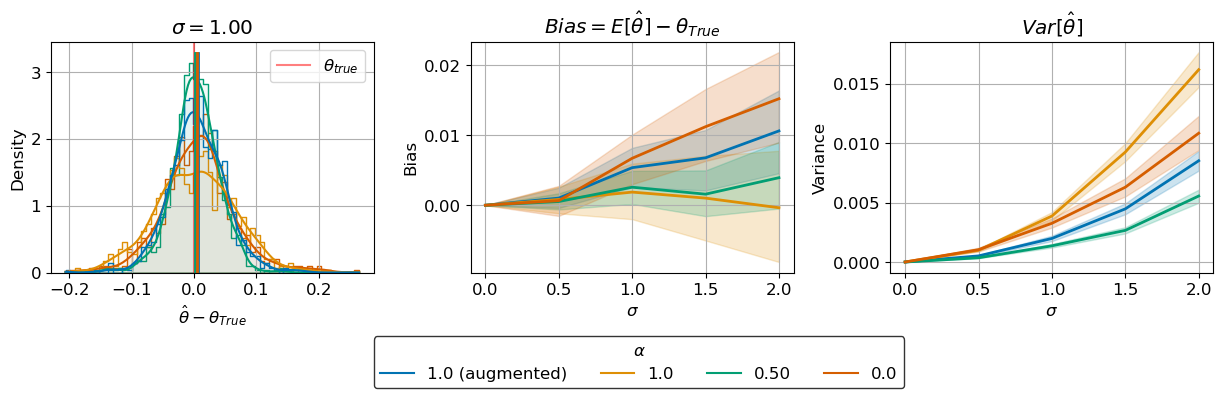}
\end{center}
\caption{The left panel shows the density distribution of $\hat{\theta} - \theta_{true}$ for a fixed $\sigma$ of 1.0. The middle panel delineates the bias of the estimator, defined as $E[\hat{\theta}] - \theta_{true}$, across varying levels of $\sigma$, and weights $\alpha \in \{0, 0.5, 1\}$, in addition to the augmented variant of the $\alpha=1.0$ model. The right panel presents the variance of the estimator, $Var[\hat{\theta}]$, as a function of $\sigma$ for the same set of $\alpha$ values. The shaded regions represent the 95\% bootstrap confidence intervals across 10 $\theta_{true}$ values, and 100 Monte-Carlo simulations.}
\label{fig:augmented}
\end{figure}

As suggested by the previous reasoning, \cref{fig:augmented} shows that indeed augmenting the training data of the baseline model reduces its variance, yet the multi-step model with $\alpha=0.5$ remains the optimal in terms of variance. Interestingly, we notice that the bias increases for the data-augmented model, this is due to the noise appearing in the inputs now, which changes the closed-form solutions derived in \cref{subsec:Uni} and the estimator is no longer unbiased. We conclude on the optimality of the multi-step even in the data augmented case.

\subsubsection{Analogy to noise reduction by averaging}
\label{appendix:uni_noise_reduction}

\begin{figure}[ht]
\begin{center}
   \includegraphics[width=.95\linewidth]{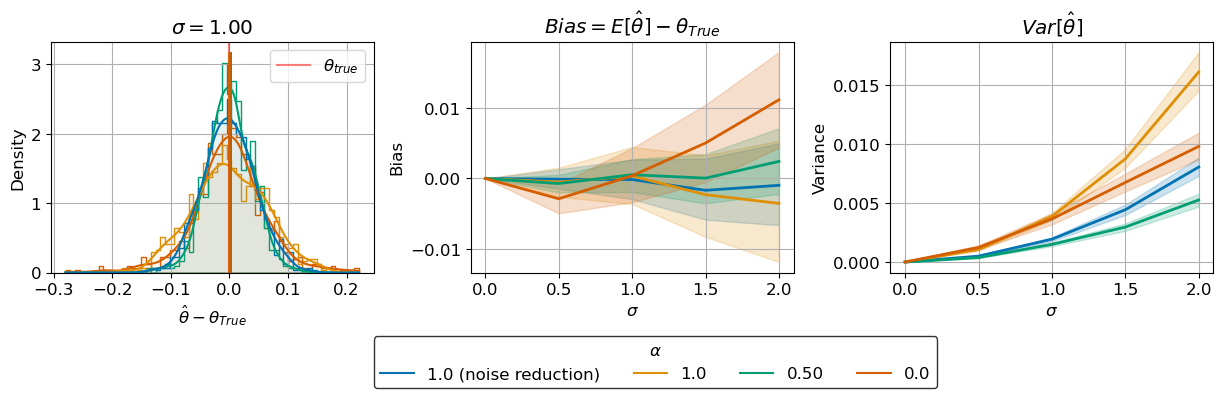}
\end{center}
\caption{Similar to \cref{fig:augmented}, in addition to the noise reduction variant of the $\alpha=1.0$ model.}
\label{fig:noise_reduction}
\end{figure}

One way to increase the signal-to-noise ratio (SNR) is \emph{averaging} \citep{Ng2010, Drongelen2007}. It consists in sampling the same noisy event $k$ times, and observing a noise reduction in the average by a factor of $\sqrt{k}$. Although we have no guarantees about the factor of noise reduction observed when we use the multi-step loss, we hypothesize that the observed improvement is similar to that of noise reduction by averaging.

To test this hypothesis, we reconduct the bias-variance analysis in \cref{subsec:Uni} and \cref{appendix:uni_data_aug} with a noise reduction variant that consists in training the $\alpha=1.0$ model (baseline) on a modified dataset: $\cD_{reduc} = \{\cS_0\}$, with the next states being $y = \{ \frac{\cO_1 + \cO'_1}{2}\}$, where $\cO_1$ and $\cO'_1$ are two realizations of the noisy next transitions. As illustrated in \cref{fig:noise_reduction}, the noise reduction approach applied to the $h=1$ model demonstrated a significant reduction in variance while maintaining unbiasedness, a result that is supported by theoretical justification. Nonetheless, it is important to note that the multi-step model with $\alpha=0.5$ continues to display a lower variance, albeit with a marginally increased bias.

\subsection{Two-parameter Non-linear system}
\label{appendix:bi}

\subsubsection{The loss landscape and approximate global minima}
\label{appendix:bi_loss_landscape}

\begin{figure}[ht]
     \centering
     \begin{subfigure}
         \centering
         \includegraphics[width=0.8\columnwidth]{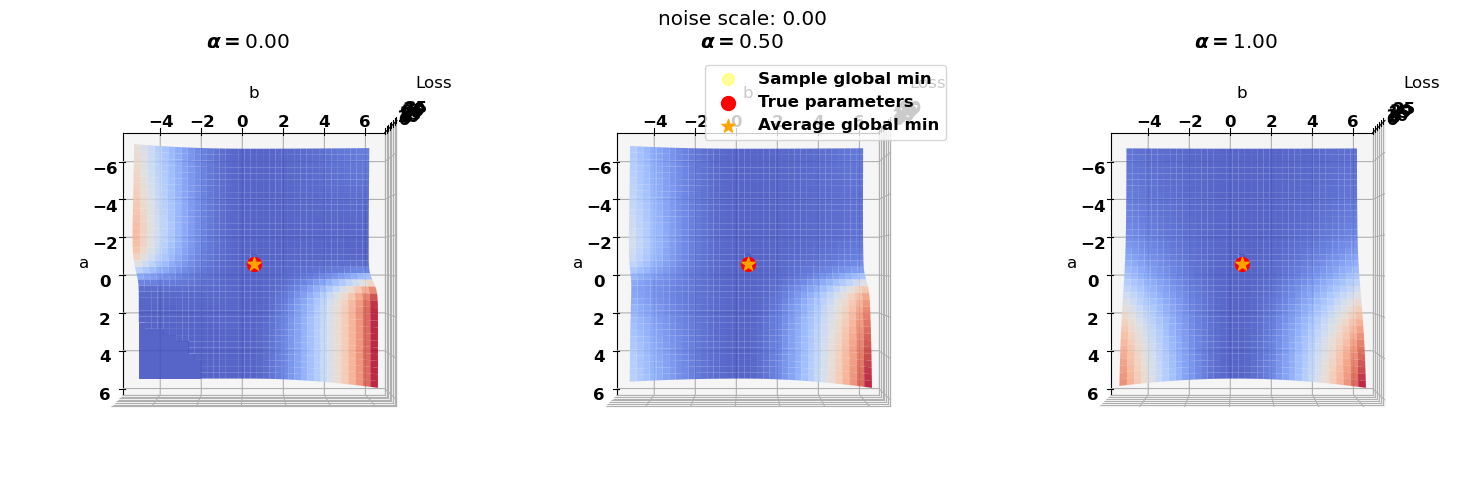}
     \end{subfigure}
     \vskip -0.4in
     \begin{subfigure}
         \centering
         \includegraphics[width=0.8\columnwidth]{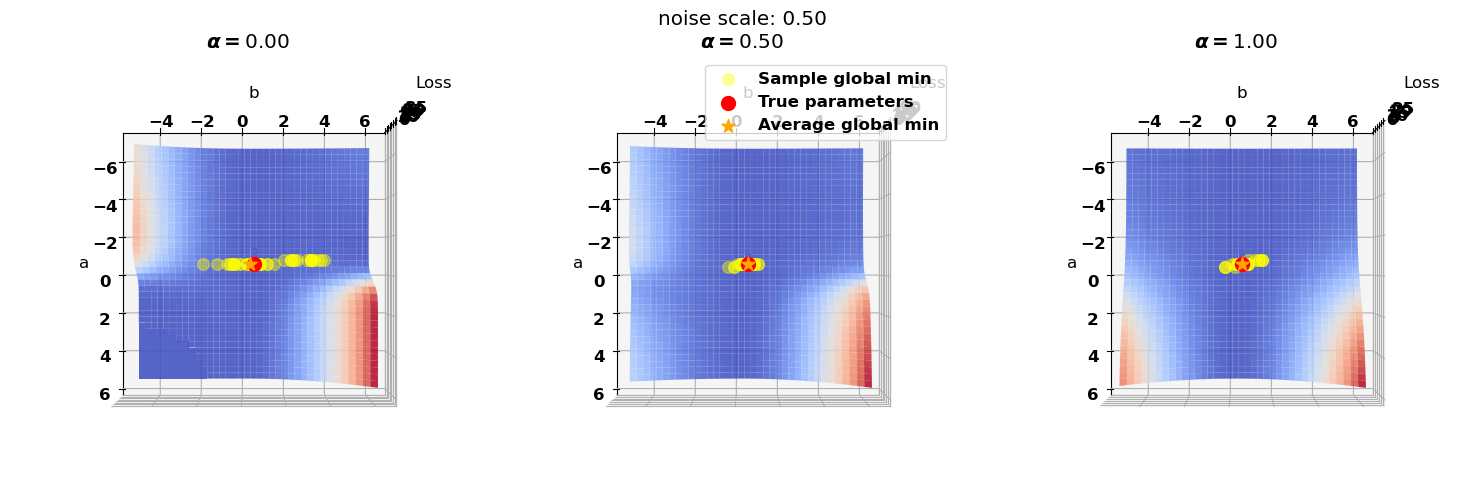}
     \end{subfigure}
     \vskip -0.4in
      \begin{subfigure}
         \centering
         \includegraphics[width=0.8\columnwidth]{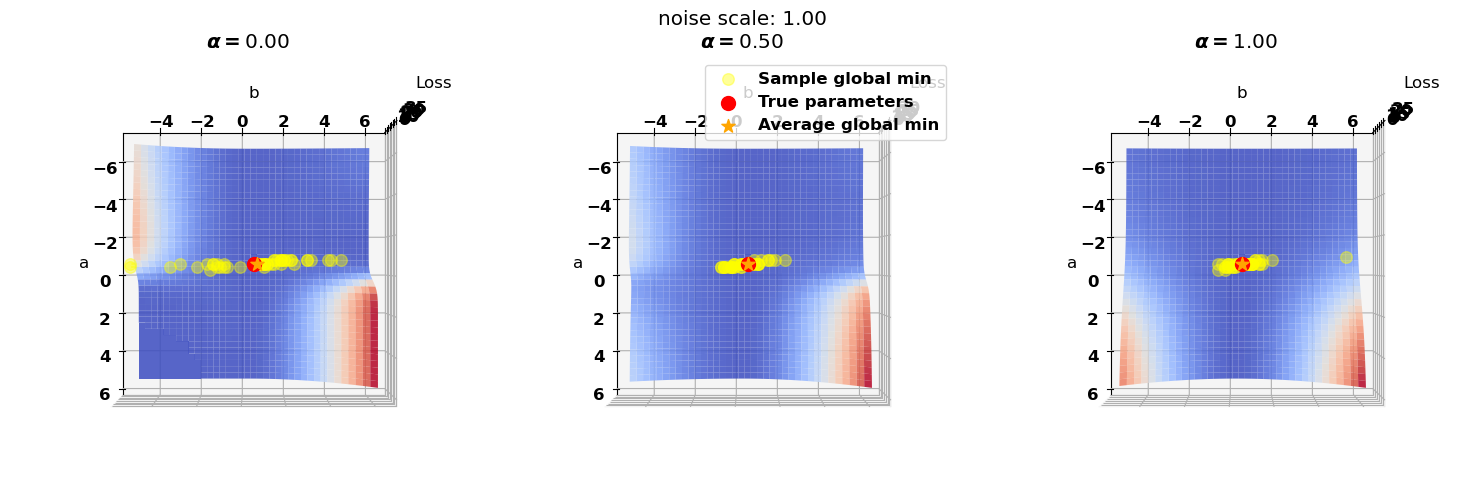}
     \end{subfigure}
     \vskip -0.4in
      \begin{subfigure}
         \centering
         \includegraphics[width=0.8\columnwidth]{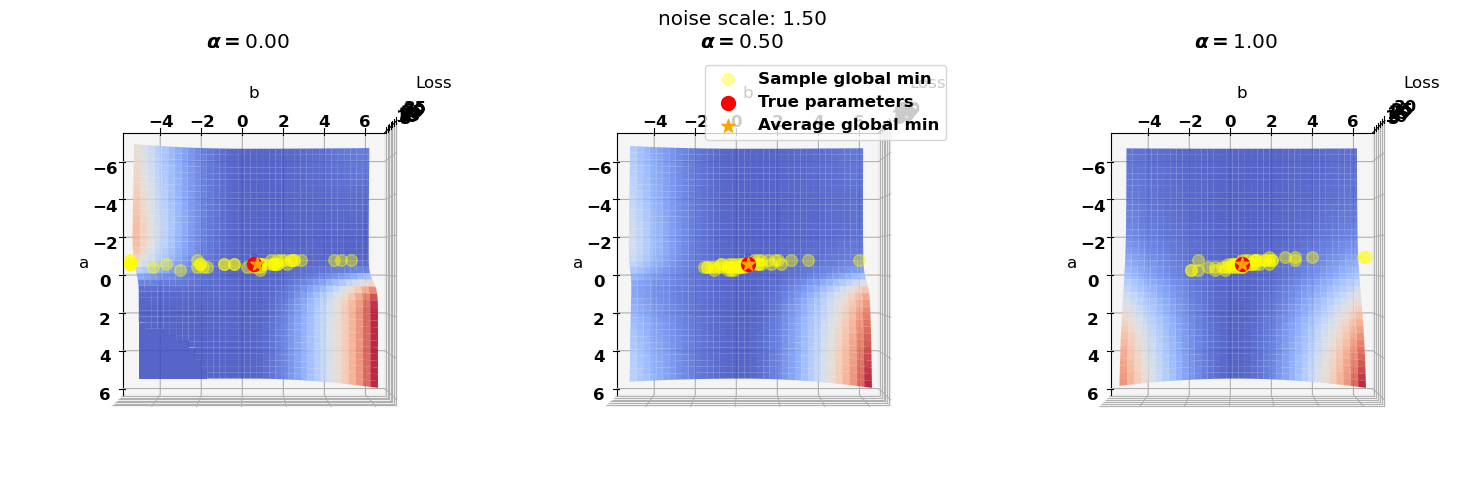}
     \end{subfigure}
     \vskip -0.4in
    \caption{This figure illustrates the impact of noise on the optimization of a multi-step MSE loss function by displaying loss landscapes at varying noise levels and alpha (\(\alpha\)) values. The true parameters (\(\theta^{true}\)) are marked with red dots. Each row represents a different level of noise, with yellow dots indicating individual instances of global minima and orange stars denoting their average.}
    \label{fig:landscape}
\end{figure}


In the non-linear case, the problem of computing the minimizers of the loss function is intractable. To build an understanding about the optimization challenges in this setup, we tried to replicate the same analysis we did in the linear case, visualizing the loss landscape and its approximate global minima \cref{fig:landscape}. 

The global minima in \cref{fig:landscape} were determined by evaluating the loss function across a finely meshed grid surrounding the true parameters $\theta^{true}$. This process was carried out for ten distinct sets of future observations at varying noise levels. The resulting data include the average loss surface, individual instances of global minima (indicated by yellow dots) and their mean (shown as an orange star). In absence of noise, all global minima are aligned with $\theta^{true}$, serving as a preliminary validation of the multi-step loss's consistency. However, as the noise level increases, the challenge of optimizing the two-step MSE loss becomes more and more apparent, evidenced by an increasing number of minimum points diverging from the true parameters. In contrast, the model with an intermediate value of $\alpha=0.5$ either matched or outperformed the baseline model ($\alpha=1$) in terms of the proximity of the global minima to the true parameters. Consequently, in this basic non-linear scenario, it is evident that the multi-step loss does not introduce significant optimization difficulties at lower $\alpha$ values.

\subsubsection{Details of the experiment in \cref{subsec:Bi}}
\label{appendix:bi_exp}

The goal of the experiment in \cref{subsec:Bi} is to assess the solutions of the multi-step MSE loss after training. Indeed, the intractability of the optimization problem forces to only analyze the solution after a gradient-based optimization procedure, all while integrating counfounding factors of this latter. The confounding factors that influence the outcome of the optimization incude: the optimizer, the initial points distribution, the learning rate, among others.

\begin{figure}[ht]
\begin{center}
   \includegraphics[width=.8\columnwidth]{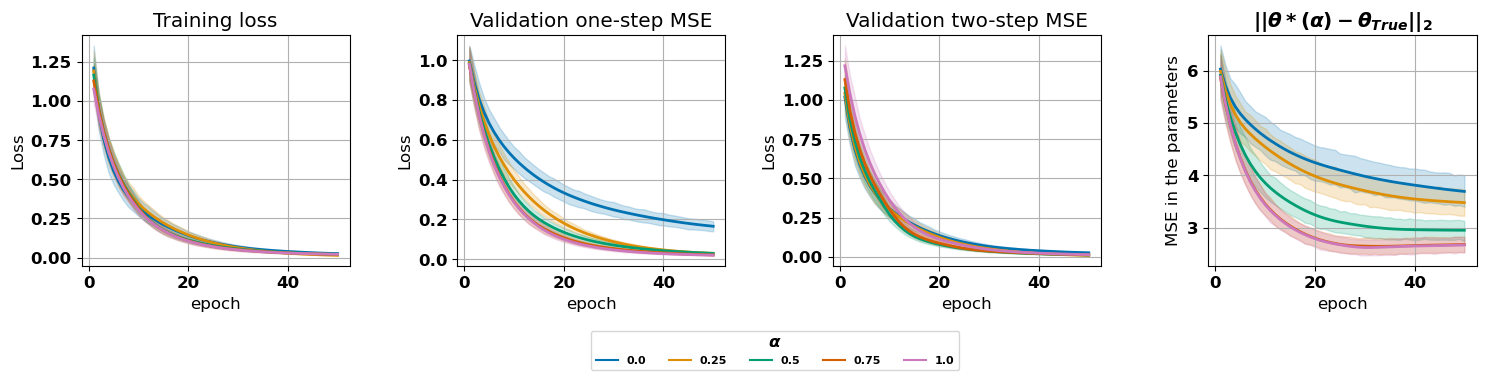}
\end{center}
\caption{The three left panels show the training, the one-step, and the two-step validation losses across 50 training epochs. The right panel show the evolution of the distance of the NN parameters to the system parameters during training.}
\label{fig:bi_nn_convergence}
\end{figure}

In this study we generate $10$ Monte-Carlo simulations for each of: $10$ randomly sampled initial points, $3$ initialization distributions (\emph{default}, \emph{uniform}, \emph{Xavier\_uniform}), $2$ optimizers (\emph{Adam}, \emph{SGD}), $3$ noise levels ($0\%$, $20\%$, and $40\%$), 5 different values of the multi-step loss parameter $\alpha \in \{0, 0.25, 0.5, 0.75, 1\}$. For each of these experiments, we train the neural network to minimize the multi-step MSE loss $L_\alpha$ for $60$ epochs. We report and analyze the validation losses and proximity to the true parameters in \cref{fig:bi_nn_ablation}, \cref{subsec:Bi}.

\cref{fig:bi_nn_convergence} shows the per-epoch training and validation loss curves, in addition to the MSE between the model parameters and the true parameters. 



\subsection{Static evaluation}
\label{appendix:static_section}

\subsubsection{$\overline{R2}(100)$ table}
\label{appendix:r2_table}

The static evaluation has been conducted by training models on the multi-step loss for different values of $h \in \{2,3,4,10\}$ and $\beta \in \{0.1, 0.3, 0.5, 0.75, 1.0, 1.5, 2.0, 3.0, 5.0, 20.0\}$. \cref{table:r2_acc} shows the corresponding test $\overline{R2}(100)$ scores and their standard deviations after the symbol $\pm$. For each horizon $h$, we show the best value of the $\beta$ parameter, as it's different across environments, datasets, and noise scales.  

\begin{table*}[h!]
\centering 
\resizebox{.999\columnwidth}{!}{
\begin{tabular}{lllllllll}
\toprule
 Environment & Dataset & Noise scale & One-step & \multicolumn{5}{c}{Multi-step} \\ \hline
 &  &  & & h=2 $[\beta]$ & h=3 $[\beta]$ & h=4 $[\beta]$ & h=10 $[\beta]$ \\ \cline{5-8}


\multirow{9}{*}{Cartpole swing-up} & \multirow{3}{*}{random} & 0 & 972 +- 4 & 975 +- 1 [0.1] & 977 +- 2 [0.3] & 980 +- 1 [0.1] & \textbf{986 +- 1 [0.75]} \\
 &  & 0.01 & 864 +- 5 & 930 +- 1 [2.0] & \textbf{950 +- 2 [2.0]} & 946 +- 3 [1.0] & \textbf{954 +- 4 [0.75]} \\ 
 &  & 0.02 & 508 +- 16 & 690 +- 15 [2.0] & 769 +- 7 [2.0] & 812 +- 7 [1.5] & \textbf{836 +- 7 [0.75]} \\ \cline{2-8} 
 
 & \multirow{3}{*}{mixed\_replay} & 0 & 812 +- 106 & 915 +- 22 [0.75] & 921 +- 32 [2.0] & 925 +- 24 [0.5] & 931 +- 20 [0.75] \\
 &  & 0.01 & 541 +- 51 & 574 +- 26 [0.1] & 659 +- 44 [3.0] & 653 +- 51 [1.5] & 679 +- 38 [0.75] \\  
 &  & 0.02 & 428 +- 9 & 335 +- 4 [2.0] & 459 +- 71 [0.75] & 481 +- 12 [1.0] & 457 +- 38 [0.75] \\ \cline{2-8} 
 
 & \multirow{3}{*}{full\_replay} & 0 & 705 +- 52 & 828 +- 6 [2.0] & 843 +- 42 [2.0] & 858 +- 32 [0.75] & \textbf{880 +- 21 [0.5]} \\
 &  & 0.01 & 714 +- 7 & 709 +- 10 [0.1] & 718 +- 6 [1.5] & 722 +- 33 [0.75] & 683 +- 32 [0.1] \\  
 &  & 0.02 & 530 +- 14 & 551 +- 34 [0.1] & 527 +- 21 [1.5] & 530 +- 46 [0.1] & 541 +- 24 [0.1] \\ \midrule 


\multirow{6}{*}{Swimmer} & \multirow{3}{*}{random} & 0 & \textbf{983 +- 1} & 974 +- 3 [0.1] & 978 +- 0 [0.1] & 975 +- 1 [0.1] & 974 +- 2 [0.1] \\
 &  & 0.01 & 934 +- 6 & 941 +- 0 [0.1] & \textbf{941 +- 2 [0.1]} & \textbf{942 +- 1 [0.5]} & \textbf{943 +- 2 [0.5]} \\  
 &  & 0.02 & 865 +- 14 & 869 +- 17 [0.1] & 880 +- 8 [0.5] & \textbf{892 +- 4 [0.5]} & \textbf{891 +- 4 [0.5]} \\ \cline{2-8} 
 
 & \multirow{3}{*}{mixed\_replay} & 0 & 609 +- 46 & 680 +- 184 [20.0] & 734 +- 18 [20.0] & \textbf{875 +- 25 [3.0]} & 735 +- 120 [0.75] \\
 &  & 0.01 & 874 +- 14 & 904 +- 16 [0.5] & 901 +- 16 [3.0] & \textbf{936 +- 5 [0.1]} & 920 +- 8 [0.75] \\  
 &  & 0.02 & 850 +- 13 & 893 +- 8 [1.0] & 878 +- 6 [0.3] & 886 +- 4 [1.0] & 883 +- 8 [0.75] \\ \midrule


\multirow{9}{*}{Halfcheetah} & \multirow{3}{*}{random} & 0 & 755 +- 3 & 746 +- 7 [0.1] & \textbf{775 +- 3 [0.1]} & 765 +- 9 [0.5] & 766 +- 5 [0.3] \\
 &  & 0.01 & 737 +- 4 & 732 +- 27 [0.1] & 756 +- 9 [0.1] & \textbf{763 +- 6 [0.1]} & 725 +- 5 [0.5] \\  
 &  & 0.02 & 704 +- 3 & 711 +- 17 [0.1] & 709 +- 23 [0.1] & 701 +- 3 [0.5] & \textbf{731 +- 0} [0.3] \\ \cline{2-8} 
 
 & \multirow{3}{*}{medium} & 0 & 516 +- 19 & 268 +- 51 [0.1] & 547 +- 10 [0.3] & 562 +- 63 [0.3] & \textbf{640 +- 13 [0.3]} \\
 &  & 0.01 & 691 +- 11 & 634 +- 20 [0.1] & 674 +- 28 [0.1] & 695 +- 3 [0.1] & \textbf{710 +- 4 [0.1]} \\  
 &  & 0.02 & \textbf{718 +- 4} & 422 +- 242 [0.1] & 675 +- 15 [0.1] & 690 +- 1 [0.1] & 702 +- 13 [0.1] \\ \cline{2-8} 
 
 & \multirow{3}{*}{medium\_replay} & 0 & 541 +- 54 & 446 +- 70 [3.0] & 589 +- 15 [0.75] & 632 +- 7 [0.75] & \textbf{771 +- 12 [0.5]} \\
 &  & 0.01 &  709 +- 12 & 725 +- 4 [0.1] & 737 +- 7 [0.1] & 737 +- 16 [0.1] & \textbf{773 +- 9 [0.3]} \\  
 &  & 0.02 &  687 +- 26 & 723 +- 16 [0.1] & 731 +- 6 [0.1] & 765 +- 18 [0.3] & \textbf{750 +- 7 [0.3]} \\  \bottomrule
\end{tabular} 
}
\caption{$\overline{R2}(100)$ for different environments, datasets, and noise scales. We highlight entries that have significantly larger score. In addition, to the mean $^pm$ standard deviation of the reported metric, the table also shows the best $\beta$ selected for each loss horizon $h$.}
\label{table:r2_acc}
\end{table*}

\subsubsection{The weight profile and the effective horizon}
\label{appendix:effective_horizon}

We can define the effective horizon $h_e$ that reflects the prediction horizon at which we effectively optimize the prediction error: given a weighted multi-step loss $L_\balpha$ with nominal horizon $h$ and weights $\balpha$, the effective horizon $h_e$ is defined as
$h_e = \sum_{i=1}^h \alpha_i \cdot i$.

For a given model trained using the multi-step loss, the optimal effective horizon is an indication of the time scale needed for optimal performance. However, models that have a different nominal horizon $h$ and the same effective horizon $h_e$ do not necessarily have the same performance. Precisely, the loss with the larger nominal horizon is setting small weights on the furthest horizons, which has a direct impact on the loss landscape and consequently on the optimization process.

\cref{table:effective_horizon} shows the optimal effective horizon $h_e$, and the exponentially parametrized weight profiles (characterized by the decay parameter $\beta$), for different values of the horizon $h$ and the noise scale $\sigma$.

\begin{table}[!htp]
\small
\caption{Best $h_e (\beta)$ values found with a grid search for each horizon and each noise scale. The values are averaged over the eight datasets.}
\label{table:effective_horizon}
\centering
\vspace{0.5em}
\begin{tabular}{l|llll}
\toprule
& \multicolumn{4}{c}{horizon $h$} \\ \midrule
 & 2 & 3 & 4 & 10 \\ \midrule
$\sigma$ (\%) & \multicolumn{4}{c}{$h_e \hspace{0.1cm} (\beta)$} \\ \hline
 0.0 & 1.30 (0.81) & 1.48 (0.45) & 1.58 (0.41) & 2.09 (0.46) \\
 0.01 & 1.26 (0.56) & 1.65 (0.76) & 1.74 (0.51) & 2.23 (0.47) \\
 0.02 & 1.36 (0.86) & 1.65 (0.72) & 2.08 (0.78) & 2.72 (0.54) \\
 0.03 & 1.33 (0.67) & 1.95 (1.21) & 2.13 (0.81) & 2.33 (0.50) \\
 0.04 & 1.40 (0.74) & 1.88 (1.01) & 2.02 (0.74) & 2.49 (0.55) \\
 0.05 & 1.49 (1.33) & 1.76 (0.83) & 2.28 (0.97) & 2.22 (0.51) \\ \bottomrule
\end{tabular}
\end{table}

The main insight of \cref{table:effective_horizon} is that regardless of the nominal horizon $h$, the effective horizon $h_e$ (and equivalently the decay parameter $\beta$) increases with the noise scale. This finding supports the idea that multi-step models are increasingly needed when incorporating information from the future is crucial to achieve noise reduction. As discussed in the previous experiment, the results are highly dependent on the task (environment/dataset), while in \cref{table:effective_horizon} we aggregate the results across tasks, and still observe the increasing trend.

Another important result highlighted in \cref{table:effective_horizon} is the upper bound on the effective horizon ($h_e$ does not go beyond $2.52$), even when the nominal horizon is large (e.g 10). This suggests that while putting weight on future horizons error does help the model, it is not beneficial to fully optimize for these horizons. Indeed, the additional components of the multi-step MSE loss act as a regularizer to the one-step loss, rather than a completely different training objective.

\subsubsection{$R2(h)$ curves}
\label{appendix:r2_curves}

In this section, we show the full $R2(h)$ curves for all environments / datasets.

\begin{figure}[ht]
     \centering
     \begin{subfigure}
         \centering
         \includegraphics[width=\columnwidth]{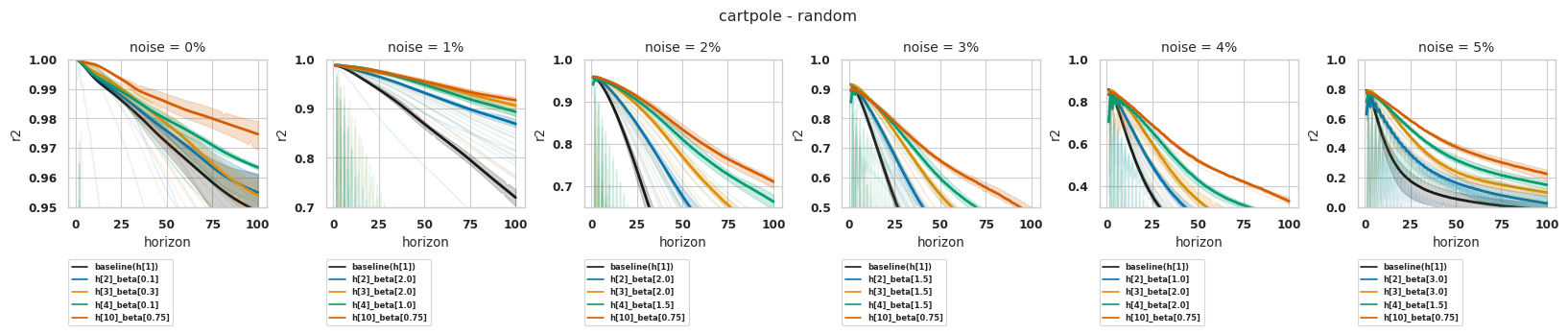}
     \end{subfigure}
     \vskip -0.4in
      \begin{subfigure}
         \centering
         \includegraphics[width=\columnwidth]{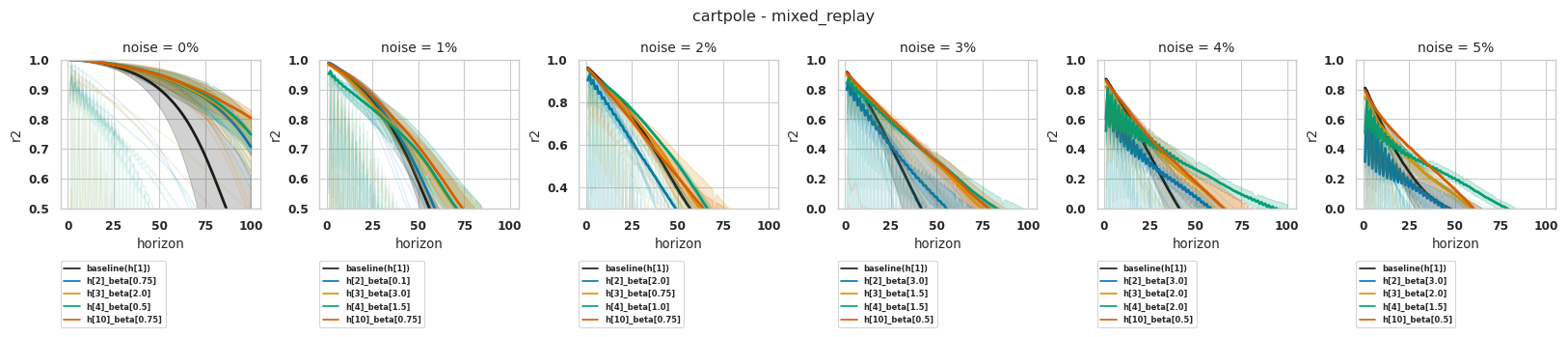}
     \end{subfigure}
     \vskip -0.4in
      \begin{subfigure}
         \centering
         \includegraphics[width=\columnwidth]{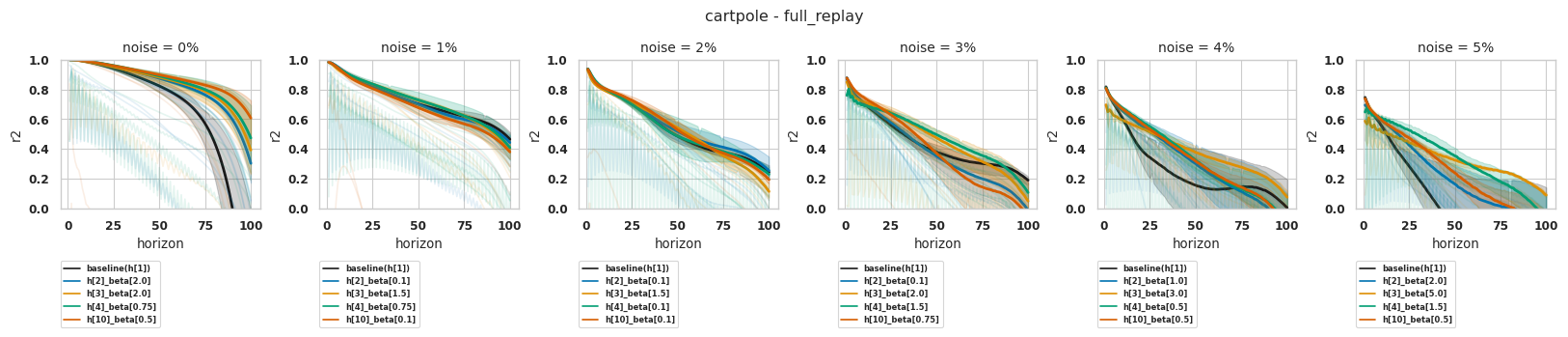}
     \end{subfigure}
    \caption{Cartpole.}
    \label{fig:r2_cartpole}
\end{figure}

\begin{figure}[ht]
     \centering
      \begin{subfigure}
         \centering
         \includegraphics[width=\columnwidth]{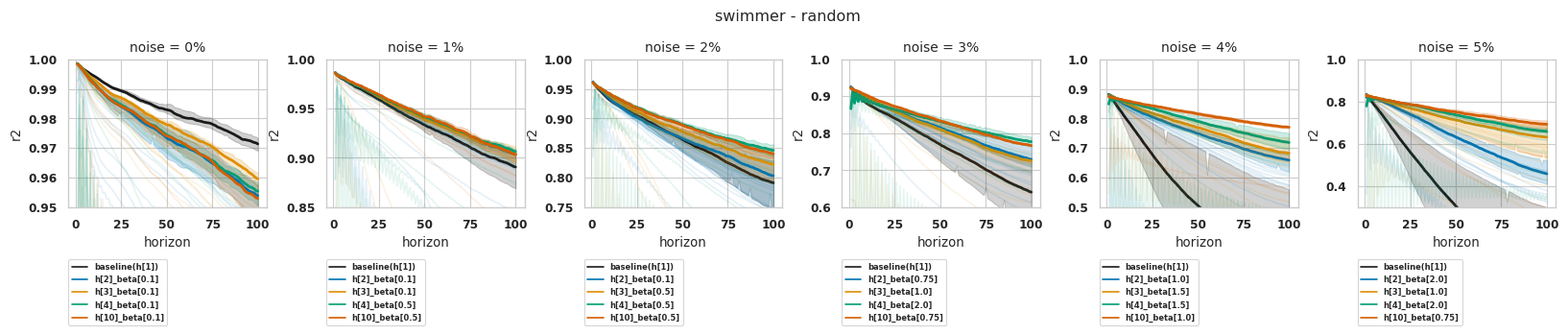}
     \end{subfigure}
     \vskip -0.4in
      \begin{subfigure}
         \centering
         \includegraphics[width=\columnwidth]{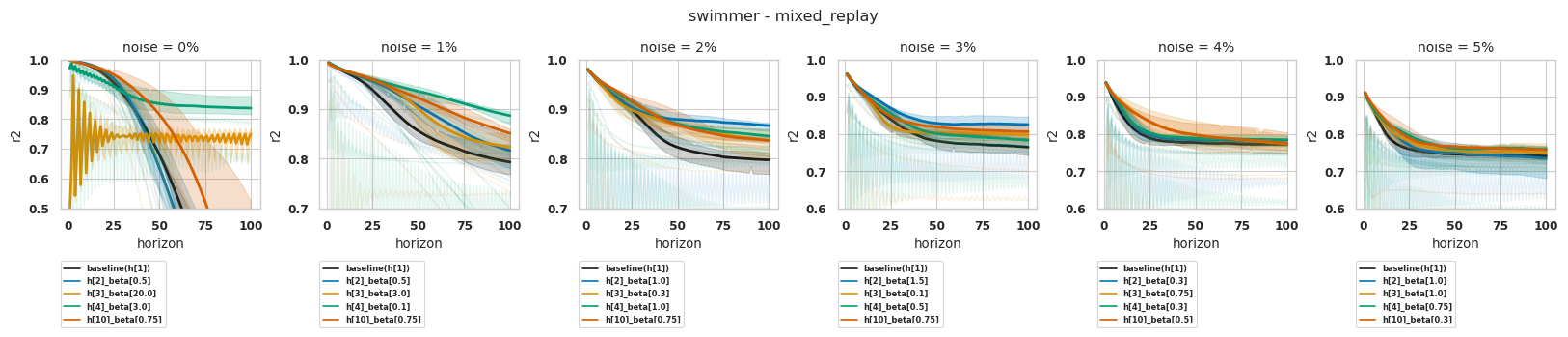}
     \end{subfigure}
    \caption{Swimmer.}
    \label{fig:r2_swimmer}
\end{figure}

\begin{figure}[ht]
     \centering
     \begin{subfigure}
         \centering
         \includegraphics[width=\columnwidth]{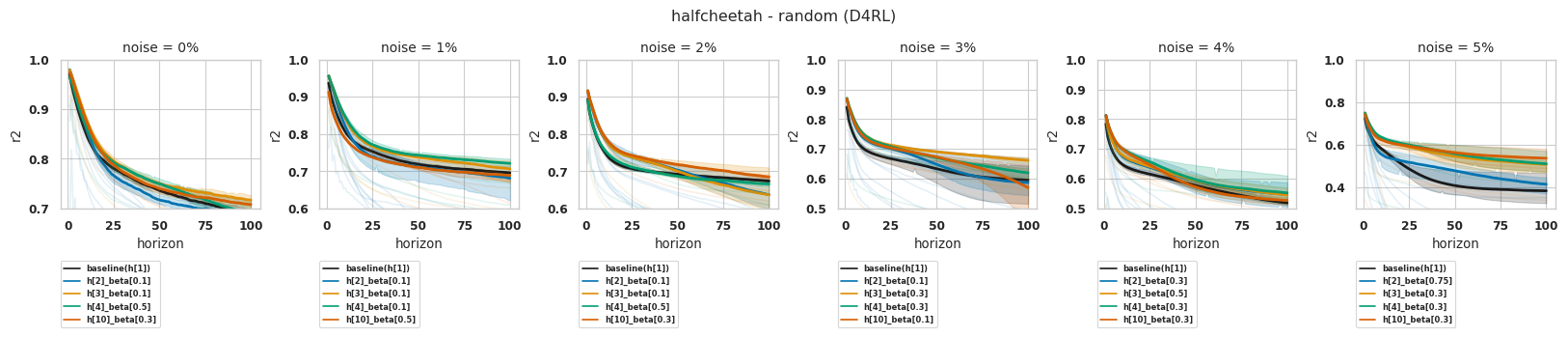}
     \end{subfigure}
     \vskip -0.4in
      \begin{subfigure}
         \centering
         \includegraphics[width=\columnwidth]{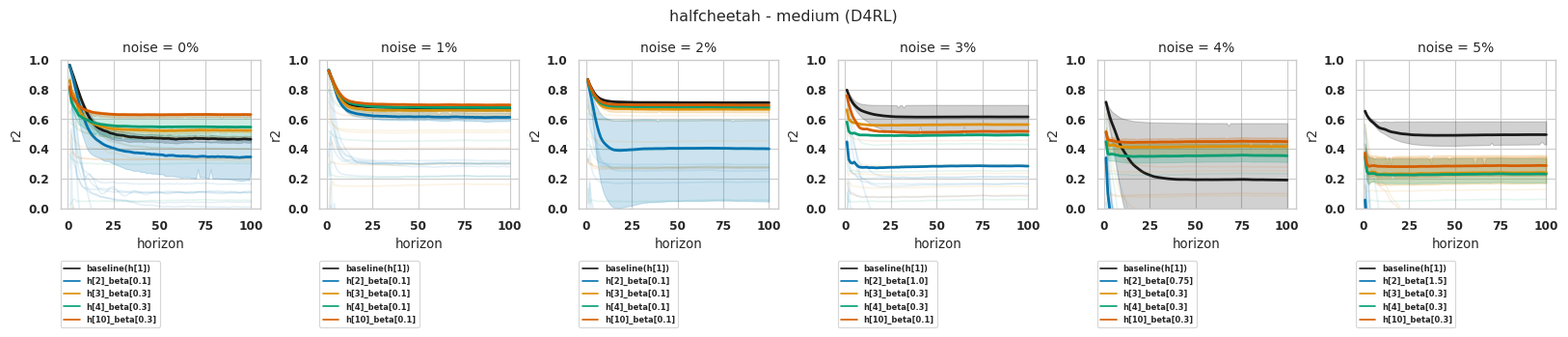}
     \end{subfigure}
     \vskip -0.4in
      \begin{subfigure}
         \centering
         \includegraphics[width=\columnwidth]{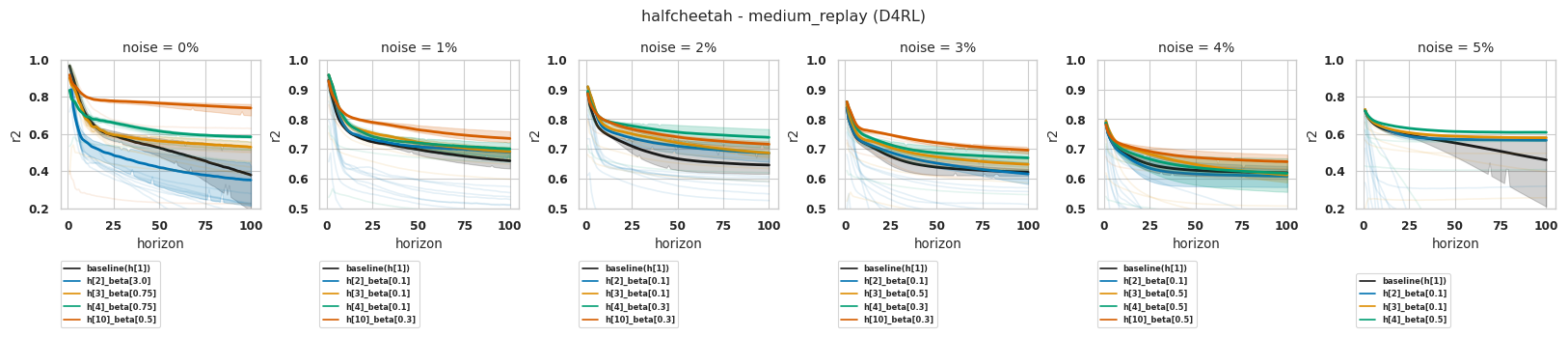}
     \end{subfigure}
     \caption{Halfcheetah.}
    \label{fig:r2_halfcheetah}
\end{figure}




\end{document}